%% file: arxiv.tex
\documentclass[lettersize,journal]{IEEEtran}
\usepackage{amsmath,amsfonts}
\usepackage{algorithmic}
\usepackage{array}
\usepackage{textcomp}
\usepackage{stfloats}
\usepackage{url}
\usepackage{verbatim}
\usepackage{graphicx}
\usepackage{amssymb}
\usepackage{amsmath}
\usepackage{graphicx}
\usepackage{pifont}
\usepackage{amssymb}
\usepackage{makecell}
\usepackage{multirow}
\usepackage{caption}
\usepackage{booktabs}
\usepackage{mathtools} 
\usepackage {bbm} 
\usepackage{varwidth}
\usepackage{wrapfig,lipsum,booktabs}
\usepackage{xspace}
\usepackage[table]{xcolor}
\usepackage{cite}
\usepackage[ruled,vlined,linesnumbered]{algorithm2e}
\usepackage[
  citebordercolor={0.2 0.5 0.9} 
]{hyperref}

\usepackage{colortbl,xcolor}
\newcommand{\method}{\textsc{VINE}\xspace}
\newcommand{\methodbold}{\textbf{\textsc{VINE}\xspace}}

\usepackage[caption=false,font=footnotesize]{subfig}

\def\BibTeX{{\rm B\kern-.05em{\sc i\kern-.025em b}\kern-.08em
    T\kern-.1667em\lower.7ex\hbox{E}\kern-.125emX}}
\usepackage{balance}
\begin{document}
\title{Hierarchical Vision Language Action Model \\
Using Success and Failure Demonstrations}
\author{Jeonguen Park$^1$, Jihwan Yoon$^1$, Byungwoo Jeon$^2$, Juhan Park$^1$, Jinwoo Shin$^2$, \\ Namhoon Cho$^3$, Kyungjae Lee$^4$, Sangdoo Yun$^5$, and Sungjoon Choi$^1$ \\
\href{https://vine-vla.github.io/}{\textcolor[rgb]{0.9,0.2,0.5}{\texttt{vine-vla.github.io}}}
\thanks{$^1$Jeongeun Park, Jiwhan Yoon, Juhan Park and Sungjoon Choi are with 
the Department of Artificial Intelligence, 
Korea University, Seoul, Korea, 
$^2$Byungwoo Jeon and Jinwoo Shin are with the Kim Jaechul Graduate School of AI, KAIST, Seoul, Korea, 
$^3$Namhoon Cho is with the Department of Aerospace Engineering, Seoul National University, Seoul, Korea,
$^4$Kyungjae Lee is with the Departmnet of Statistics, Korea University, Seoul, Korea, 
$^5$Sangdoo Yun is with the NAVER AI Lab, Seongnam, Korea}}
\markboth{Preprint Version. Under Review. }{}

\maketitle

\begin{abstract}
Prior Vision–Language–Action (VLA) models are typically trained on teleoperated \emph{successful} demonstrations, while discarding numerous \emph{failed} attempts that occur naturally during data collection. 
However, these failures encode where and how policies can be fragile, information that can be exploited to improve robustness. 
We address this problem by leveraging mixed-quality datasets to learn \emph{failure-aware reasoning} at planning time. 
We introduce \methodbold, a hierarchical vision-language-action model that separates high-level reasoning (System~2) from low-level control (System~1) under a hierarchical reinforcement learning formalism, making failures usable as a structured learning signal rather than noisy supervision. System~2 performs feasibility-guided tree search over a 2D scene-graph abstraction: it proposes subgoal transitions, predicts success probabilities from both successes and failures, and prunes brittle branches before execution, effectively casting plan evaluation as feasibility scoring. The selected subgoal sequence is then passed to System~1, which executes low-level actions without modifying the agent’s core skills. Trained entirely from offline teleoperation data, \methodbold~integrates negative experience directly into the decision loop. Across challenging manipulation tasks, this approach consistently improves success rates and robustness, demonstrating that failure data is an essential resource for converting the broad competence of VLAs into robust execution.
\end{abstract}

\input{sections/1.introduction}
\input{sections/2.related_work}

\input{sections/3.formulation}

\input{sections/4.method}
\input{sections/5.experiments}

\input{sections/6.conclusion}

\bibliographystyle{unsrt} 
\bibliography{reference} 

\input{sections/appendix}
\end{document}

%% file: sections/1.introduction.tex
\section{Introduction}



In this work, we address the problem of leveraging mixed-quality datasets~\cite{8626460, hejna2024remix} to train a Vision-Language-Action model (VLA)~\cite {bjorck2025gr00t,black2024pi_0,rt_x_2023,kim2024openvla}. Mixed-quality demonstrations, combining successful and failed trials, provide complementary supervision: successes chart feasible progress, while failures supply counterexamples that identify brittle behaviors and sharpen decision boundaries~\cite{8626460,hoang2024sprinql}.
Within VLA policies that map language and vision to actions, the prior work shows that training on such mixed-quality data improves robustness~\cite{black2024pi_0}.
Building on this insight, we propose a VLA framework that learns \emph{failure-aware reasoning} as an explicit planning mechanism: it estimates feasibility from both successes and failures to score candidate plans and prioritize high-feasibility paths \cite{ichter2022do}, suppressing brittle branches before execution.



We hypothesize that by learning not only what works but also what fails, an agent can better anticipate risky outcomes. Failure data~\cite{liu2023reflect, diehl2022did,8626460} is an often overlooked yet essential element in VLA training. Most robot data for post-training VLAs come from offline trajectories collected via human teleoperation~\cite{kim2024openvla, bjorck2025gr00t, black2024pi_0}. Although these collections naturally include failed attempts, e.g., unstable grasps and collisions, such traces are typically discarded as noise. However, they encode information about infeasible transitions and unsuccessful behaviors, making them a vital resource for improving robustness.


The challenge lies in effectively integrating this failure signal from existing offline datasets. In Imitation Learning (IL), while explicitly penalizing failure-prone transitions is possible, carefully tuning these penalties to avoid distorting the learned policy is a non-trivial challenge~\cite{8626460,belkhale2023data,hejna2024remix}. On the other hand, Reinforcement Learning (RL)~\cite{sutton1998introduction, kaelbling1996reinforcement} offers a natural framework to handle failure data through a reward signal. To utilize this property, we formulate the overall system as \emph{Hierarchical Reinforcement Learning} (HRL)~\cite{sutton1999between}. 
We introduce \methodbold, \textbf{V}ision–\textbf{L}anguage–\textbf{A}ction model \textbf{I}ntegrating \textbf{N}egative \textbf{E}xperience, a hierarchical VLA framework that integrates negative experience built upon pre-trained VLA (i.e., $\pi_0$~\cite{black2024pi_0}), and is formulated through HRL.
Our approach separates high-level planning from low-level control, and injects failure supervision solely offline to train the planner’s value modules without any online rollouts.


This HRL-inspired separation is embodied in our method's hierarchical system~\cite{shi2025hi, bjorck2025gr00t} architecture, inspired by Kahneman's cognitive process~\cite{kahneman2011thinking}. 
The high-level System 2 is responsible for reasoning and planning. 
We cast planning as a feasibility-guided tree search, scoring each candidate transition by its predicted success probability. In particular, we construct a tree of a 2D scene graph that abstracts world states (nodes), where subgoals represent the transitions (edges) connecting them. Crucially, System 2 is trained with both success and failure data to learn a feasibility score for each node, generate options, or generate an estimated next state representation. This allows the planner to conduct failure-aware reasoning, which preemptively identifies and avoids paths likely to fail. The optimal sequence of sub-goals and predicted next scene graph is then passed to the low-level policy, System 1, for execution. This design cleanly slots the learned feasibility signal into the high-level reasoning loop, enhancing robustness without altering the agent's fundamental skills.

Our contributions are summarized as follows:
(1) We leverage offline failure data to train the System~2 model and integrate it into a tree-based planner that scores each candidate step by its predicted success probability.
(2) We introduce a hierarchical VLA framework, grounded in HRL, that separates feasibility-aware high-level planning from low-level action execution.
(3) We present a comprehensive empirical evaluation showing significant gains in success rate and robustness over a strong VLA baseline on manipulation tasks.

%% file: sections/2.related_work.tex
\section{Related Work}
\subsection{Vision--Language--Action Models}
VLA models unify perception, language, and action through end-to-end learning~\cite{brohan2022rt, rt2_2023, rt_x_2023, octo_2023, wen2025diffusionvla, lbmtri2025}. 
RT-1~\cite{brohan2022rt} showed that Transformers can learn manipulation from large single-platform data; RT-2~\cite{rt2_2023} extended this by adapting web-scale vision-language models for control, transferring high-level semantics to actions. The creation of the Open-X Embodiment dataset~\cite{rt_x_2023} was a critical inflection point, aggregating data from dozens of different robots and institutions. Leveraging this massive, multi-embodiment dataset, models like RT-X~\cite{rt_x_2023}, and OpenVLA~\cite{kim2024openvla} emerged as general-purpose ``foundation models'' for robotics. However, such models employ autoregressive discretization to represent actions, limiting their ability to handle high-frequency dexterous tasks.  

While initial VLAs established the power of large-scale pretraining, more recent work has focused on improving action generation for high-frequency and dexterous tasks.
$\pi_{0}$~\cite{black2024pi_0} and $\pi_{0.5}$\cite{pi_05_2025} with PaliGemma~\cite{beyer2024paligemma} backbones for continuous actions with flow matching~\cite{lipman2023flow} objective with action chunks~\cite{zhao2023learning}, allowing high-frequency control. 
GR00T~\cite{bjorck2025gr00t} is trained on heterogeneous real robots, human-video, and synthetic data to achieve strong cross-embodiment dexterous control.
Recent work augments VLAs with explicit reasoning to boost generalization and interpretability, such as predicting textual traces~\cite{zawalski2024robotic}, next images~\cite{zhao2025cot}, or affordance~\cite{li2024improving}. 
Among these, ThinkAct~\cite{huang2025thinkact} is trained via reinforcement learning for explicit reasoning with action-aligned visual feedback.
In contrast, we introduce explicit failure modeling: a reach–avoid feasibility value learned from both successes and failures and conditioned on predicted successor states. These failure-aware estimates guide look-ahead search to prune unsafe branches and explicitly select the more feasible path.


\subsection{Hierarchical System using Language Models}
A parallel line of research couples large language models with low-level controllers, where the large (vision) language model is used primarily for symbolic planning or code generation. Early examples include Code as Policies~\cite{liang2022code} and ProgPrompt~\cite{10161317}, which translate natural language instructions into structured robot code, while ReAct~\cite{yao2023react}-style prompting leverages reasoning traces to interleave planning and action selection. 
However, these standard approaches remain limited in their ability to deeply understand the dynamics of the environment, anticipate failures, or verify the feasibility of proposed actions.

Building on this dual-system idea, recent work jointly trains both levels of the hierarchy.
Hi Robot~\cite{shi2025hi} uses a high-level VLM to decompose complex instructions into atomic language commands for a low-level policy. HAMSTER~\cite{li2025hamster} uses a high-level model to predict a 2D end-effector path, which guides a 3D-aware low-level controller and enables generalization from off-domain data. 
MolmoAct~\cite{lee2025molmoact} predicts 2D visual trajectories and depth-aware tokens for interpretable, user-steerable manipulation.
Yet, these models largely omit explicit reasoning to anticipate failures or verify feasibility, motivating failure-aware VLAs. 

In this work, we adopt a dual system and jointly train both levels: System~2 for reasoning and System~1 for control, on a shared VLA backbone with failure supervision, naturally connecting to affordance-guided selection as in SayCan~\cite{ichter2022do}. This approach evaluates multiple candidate subgoals and choose those with high affordance scores learned from robot trajectories. Rather than using failures only as an offline regularizer or a global affordance prior, we use tree-based look-ahead planning and score each candidate node with a success probability, while using the same language backbone also to produce subgoal-conditioned actions.

\subsection{Tree Search}
%
Tree search is a foundational technique in robotic planning for exploring sequences of actions and their potential outcomes~\cite{9140424, labbe2020monte,4084578, 1492476}. This principle of structured exploration has been recently adapted for language models in Tree-of-Thoughts (ToT) prompting, which casts complex reasoning as a search over branching thought states~\cite{yao2023tree}. In practice, ToT commonly leverages Monte Carlo Tree Search (MCTS)~\cite{zhao2023large, zhang2024restmcts,gao2025interpretable,chi-etal-2025-thoughtsculpt}, with nodes representing partial reasoning states and rollouts used to estimate success. Recent work emphasizes the importance of value guidance; for instance, \cite{liu2024dont} shows that a PPO-trained value model can guide MCTS to avoid prematurely pruning useful candidates. For embodied agents, ToT enables the evaluation of alternative futures to select safer actions, but its effectiveness hinges on well-calibrated value estimates that success-only data cannot provide. 
Failure trajectories supply the essential negative evidence needed to distinguish feasible from infeasible branches, making them critical for more accurate value learning.

%% file: sections/3.formulation.tex
\section{Problem Formulation}
\begin{figure}[!t]
    \vspace{-2em}
    \captionsetup{type=figure}
    \includegraphics[width=0.99\linewidth]{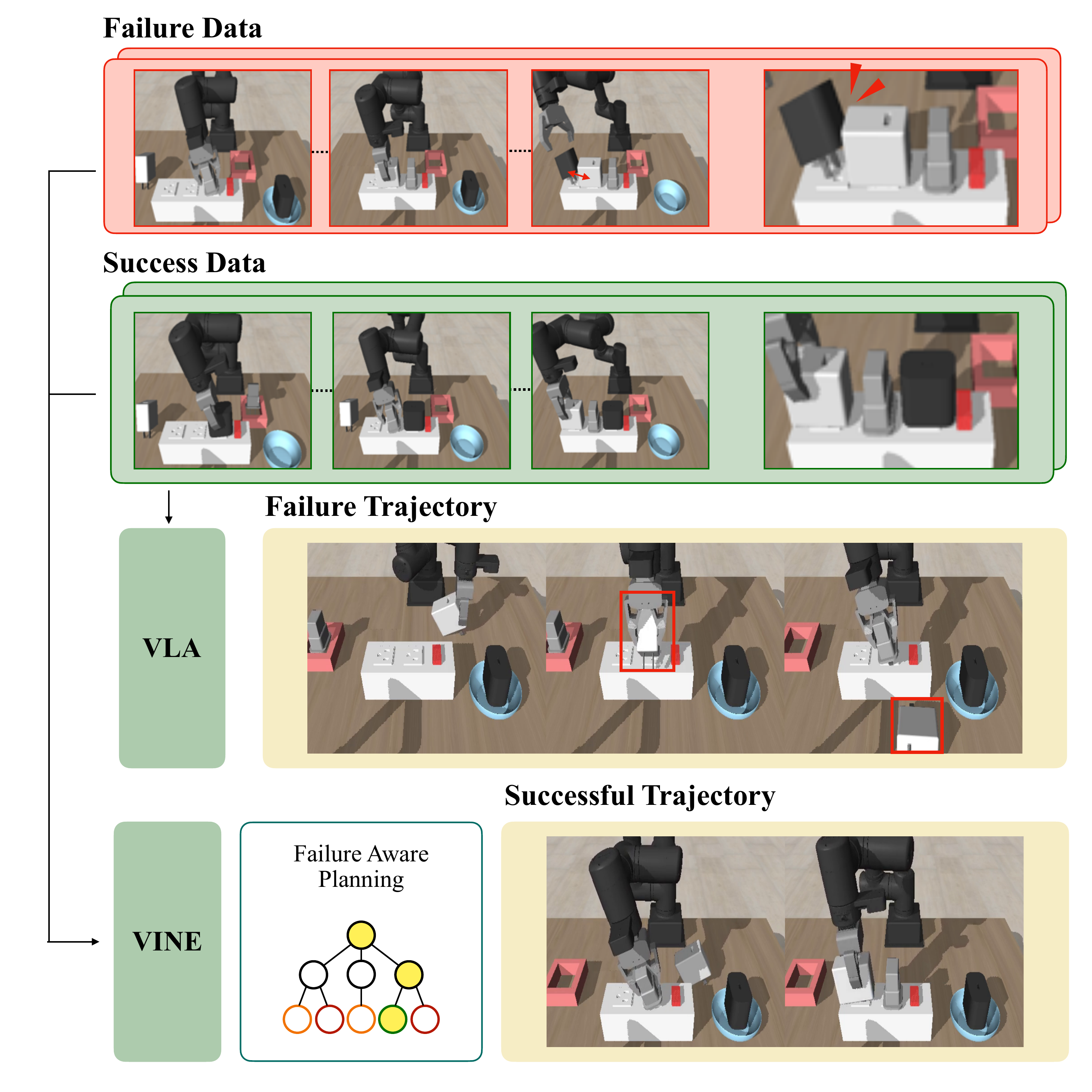}
    \caption{Standard VLAs trained only on success data may produce infeasible trajectories. VINE leverages both success and failure trajectories with a failure-aware reasoning planner, yielding more robust executions.}
    \label{fig:intro}
\end{figure}

In this section, we formalize instruction-conditioned manipulation with a reach–avoid objective from mixed-quality demonstrations (successes and failures) as shown in Figure~\ref{fig:intro}. We use a dual-system hierarchy: \textbf{System~2} performs a look-ahead tree search over candidate edges with a failure-aware feasibility/value estimator, and \textbf{System~1} executes the selected path with closed-loop control until termination. Each edge is grounded as an option, inducing an Semi-Markov decision process (SMDP) over abstract states; the following paragraphs specify the states, options, termination, and success/failure conditions.

\subsection{Hierarchical VLA}
We formalize a \emph{hierarchical VLA} framework where \textbf{System~2} performs high-level reasoning and planning, while \textbf{System~1} executes each selected plan through low-level control actions.
Most VLA policies are trained via imitation learning (IL), which is data-efficient and stable with strong demonstrations, but provides only limited leverage for explicitly using failure data~\cite{8626460,hejna2024remix,belkhale2023data}. 
Viewed through the lens of reinforcement learning (RL)~\cite{sutton1998introduction,kaelbling1996reinforcement}, incorporating both successes and failures becomes natural via reward-based learning, motivating a hierarchical reinforcement learning (HRL) formulation~\cite{sutton1999between}.
Accordingly, we adopt HRL as the formal scaffold: System~2 plays the role of the meta-controller and employs \emph{model-based tree search} with brief look-ahead over candidate subgoal transitions to estimate failure-aware value and choose a plan, while System~1 focuses on robust execution of the chosen subgoal sequence. 

\subsubsection{Data Assumption and Objective}
Our dataset $\mathcal D$ comprises teleoperated trajectories labeled as success (target state reached) or failure (instruction violation or irrecoverable state). We use a sparse terminal reward ($r_t{=}1$ on success, $0$ otherwise) and learn from tuples $(s_t,a_t,\ell,r_t)$. 
 Given $(s_0,\ell)$ and candidate plans $\mathcal{T}$ from a shallow look-ahead, \textbf{System~2} chooses $\tau^\star=\arg\max_{\tau\in\mathcal{T}} \widehat{\Pr}[\text{success}\mid \tau,s_0,\ell]$, and \textbf{System~1} executes its chosen path to termination.

\subsubsection{Prerequisites for Tree Search and Execution}
To enable hierarchical tree search, System~2 requires three components: 
(i) a world model to predict hypothetical successors offline; 
(ii) a candidate-proposal mechanism that expands only instruction-consistent edges, controlling the branching factor; and 
(iii) a failure-aware value estimator to score nodes and back up returns. 
System~1 employs an option-conditioned, closed-loop policy to execute each selected edge and a termination detector to signal arrival at the successor state.

\subsubsection{Bridge to HRL}
Interpreted in HRL~\cite{sutton1999between} terms, the chosen route can be viewed as a sequence of \emph{options}. System~2 plays the role of a meta–controller that chooses among high-level \emph{options} aligned with edges in our abstract graph. Each edge corresponds to an option with three parts: an initiation set (states consistent with the current node), an intra-option policy carried out by System~1 (a low-level controller conditioned on the selected options), and a termination rule that triggers when the successor node is reached or a failure condition is met. Because options unfold over a variable number of primitive steps, the induced high-level process can be viewed as semi-Markov decision process (SMDP).
We formalize this connection next.

\subsection{SMDP Formulation}
Let the low-level interaction of the robot be an Markov decision process
\(\mathcal{M}=(\mathcal{S},\mathcal{A},T,\rho)\) under an instruction \(\ell\),
with states \(s\in\mathcal{S}\) including images and proprioception observations, actions \(a\in\mathcal{A}\), transition kernel \(T(\cdot\mid s,a)\),
and initial distribution \(\rho\).
A \emph{node} is a compact abstraction of the robot/world state:
\(n=\phi(s)\in\mathcal{N}\), where \(\phi:\mathcal{S}\!\to\!\mathcal{N}\) extracts scene/plan tokens, spatial relations, etc.
An \emph{edge} \(e\in\mathcal{E}\) is a temporally extended subgoal executed by a closed-loop controller (System~1) until a termination condition is met. 
At \emph{decision epochs} (the sequence of nodes \((n_k)\)), the induced process over \(\mathcal{N}\) is a semi-Markov decision process (SMDP). In this view, each transition between nodes (an edge) can be naturally modeled as an \emph{option} in the SMDP framework, 
linking symbolic reasoning with executable policies. 

We distinguish between the \emph{symbolic} form of an edge and its
\emph{option} grounding.
An edge $e \in \mathcal{E}$ is a subgoal token (e.g., ``pick up spoon''),
which specifies the intended transition at the abstract reasoning level.
For execution, each edge $e$ is associated with an option
\[
o_e \;=\; \bigl(I_e^{N},\,\pi_e,\,\beta_e,\,\Pi_e\bigr),
\]
where
\emph{(i) Initiation set)} $I_e^{N}\subseteq\mathcal{N}$ specifies the nodes at which \(e\) is admissible.
Execution at node \(n\) is feasible only if \(n\in I_e^{N}\),
\emph{(ii) Intra-option policy)} $\pi_e(a\mid s,\ell)$,
\emph{(iii) Termination rule)} $\beta_e:\mathcal{S}\to[0,1]$,
and \emph{(iv) Projection)} $\Pi_e:\mathcal{S}\to\mathcal{N}$.
Thus, $e$ denotes the symbolic subgoal, while $o_e$ denotes its formal,
executable grounding as an option.

\subsection{Option-Level Representation: Nodes, Edges, and Data}
The option view above specifies \emph{what} is executed at the high level; next we specify \emph{how} these objects are realized from data. 
Concretely, mixed–quality demonstrations provide supervision to learn (i) the abstraction that maps raw states to nodes, enabling edges to be instantiated as options, and (ii) the signals needed to train the world model, proposal, policy, and termination components. 

\subsubsection{Nodes as Abstract State Representations}
A node $n_k = \phi(s_{t_k})$ denotes an abstract, grounded representation of the world at keyframe time $t_k$. We instantiate each node as a 2D scene graph~\cite{ni2024grid, rana23a,11024207} to encode not only objects but also their spatial and semantic relations. This representation lets the planner reason about consequences: by inferring how a proposed edge will alter relations in the current scene graph, it can anticipate the next abstract state $n_{k+1}$. Concretely, $\phi$ is implemented via a two-stage pipeline: an object detector (e.g., Grounding~DINO~\cite{liu2024grounding}) proposes candidate boxes, and a vision–language model (e.g., Gemini-2.5-Flash~\cite{comanici2025gemini}) filters them and predicts the relational structure, yielding a scene graph serialized in text.

\subsubsection{Edges as Transitions Between Nodes}
Each transition is treated as a subgoal, an \emph{edge} $e_k$, executed over $[t_k, t_{k+1})$. Keyframes $\{t_k\}$ are detected from consistent gripper-state events (e.g., \texttt{closed}$\!\to$\texttt{open} ending \texttt{grasp}/\texttt{insert}; \texttt{open}$\!\to$\texttt{open} after \texttt{place}/\texttt{push}). With this view, planning reduces to explicit edge selection: choosing an edge commits to a specific state transition, which is verifiable at the subsequent keyframe.

\subsubsection{Constructing Datasets for Hierarchical Learning}
From this decomposition, we build $\mathcal{D}_{\text{sys2}}$ for the System~2 and $\mathcal{D}_{\text{sys1}}$ for the System~1. $\mathcal{D}_{\text{sys2}}$ contains tuples $(n_k,e_k,n_{k+1},z_{0:k-1},\ell,s_0,r_k)$ with history context  $z_{0:k-1}=[n_0,e_0,\dots,n_{k-1}]$ and sparse rewards ($r_k{=}0$ for $k{<}K$, $r_K{\in}\{0,1\}$). $\mathcal{D}_{\text{sys1}}$ uses continuous action data conditioned on $(e_k,n_{k+1},\ell)$, i.e., $(e_k,n_{k+1},s_t,a_t,\ell)$, so that the low-level policy is guided by the \emph{intended transition} (edge) and the \emph{predicted next state} (node). This tight coupling, edge-shaped transitions as node transitions, and nodes as predictive, goal-like abstractions, enables tractable planning over feasible paths directly grounded in demonstrations.

\subsection{Option-Level Feasible Decision Making Formulation}
Conceptually, our option-level formulation is semantically similar to prior reach–avoid frameworks~\cite{margellos2011hamilton,fisac2015reach}: by identifying reach and avoid sets from mixed success/failure data and delineating the feasible decision space.
The process terminates upon first entry into either a target set $\mathcal G$ (the success or goal set to be reached) or a failure set $\mathcal F$ (the unsafe set to be avoided), with $\mathcal G\cap\mathcal F=\varnothing$. 

Formally, define the hitting times
\begin{align*}
    \tau_{\mathcal G} &\coloneqq \inf\{t\ge0:\ s_t\in\mathcal G\},\\
\tau_{\mathcal F} &\coloneqq \inf\{t\ge0:\ s_t\in\mathcal F\}, \\
\tau &\coloneqq \tau_{\mathcal G}\wedge \tau_{\mathcal F}.
\end{align*}
An episode terminates at $\tau$ with outcome $s_\tau\in\mathcal G\cup\mathcal F$.

The set $\mathcal G$ is a target set defined by design: the task is considered successful as soon as the process reaches any state in $\mathcal G$, and the episode is terminated at that point. In particular, it suffices that there exists at least one option or policy that enables reaching $\mathcal G$.  
In contrast, the failure set $\mathcal F$ is a critical set closed under options: once the process enters $\mathcal F$, no admissible option can lead it outside. Formally,
\[
\forall\, s\in\mathcal F,\ \forall\, o\ \text{admissible at }s:\quad
\operatorname{supp} P_o(\cdot\mid s)\subseteq \mathcal F,
\]
where $P_o(\cdot\mid s)$ denotes the state distribution upon termination of option $o$ starting from $s$. Thus $\mathcal F$ is absorbing at the option level.  
This induces a sparse entrance reward
$r_t\coloneqq\mathbf 1\{s_{t+1}\in\mathcal G\}$,
so that the undiscounted return
$R=\sum_{t=0}^{\tau-1}r_t$
equals $1$ if $\tau_{\mathcal G}<\tau_{\mathcal F}$ and $0$ otherwise.

\subsection{Value as Success Probability}
We score each search node by its \emph{reach--avoid success probability}. Concretely, we learn a value function $V(n)$ that estimates the probability of hitting the goal set $\mathcal G$ before the failure set $\mathcal F$, which lets us exploit both successful and failed trajectories. This value–probability equivalence is stated below.

\newtheorem{proposition}{Proposition}
\begin{proposition}[First-exit value equals reach–avoid success probability]
Assume the tree search process (node and edges) is Markov, and the sets
$\mathcal G,\mathcal F$ are absorbing (first visit terminates). Then the value function equals the probability of reaching $\mathcal G$ before $\mathcal F$, where $\tau$ is a hitting times:
\begin{equation}
V(n)\;=\;\Pr\!\big(\tau_{\mathcal G}<\tau_{\mathcal F}\,\big|\,n\big).
\label{eq:proposition}
\end{equation}
Consequently, high–value nodes approximate the probabilistic reachable set to $\mathcal G$.
\label{sec:proposition}
\end{proposition}
\noindent\emph{Proof sketch.} 
Under the first-exit formulation with absorbing $\mathcal G_N,\mathcal F_N$, the return equals the success indicator; hence $V^\mu=\mathbb E[R]=\Pr(\tau_{\mathcal G}<\tau_{\mathcal F}\mid\cdot)$. See Appendix for detailed derivation.

%% file: sections/4.method.tex
\section{Proposed Method}

\begin{figure*}
    \centering
    \includegraphics[width=0.85\linewidth]{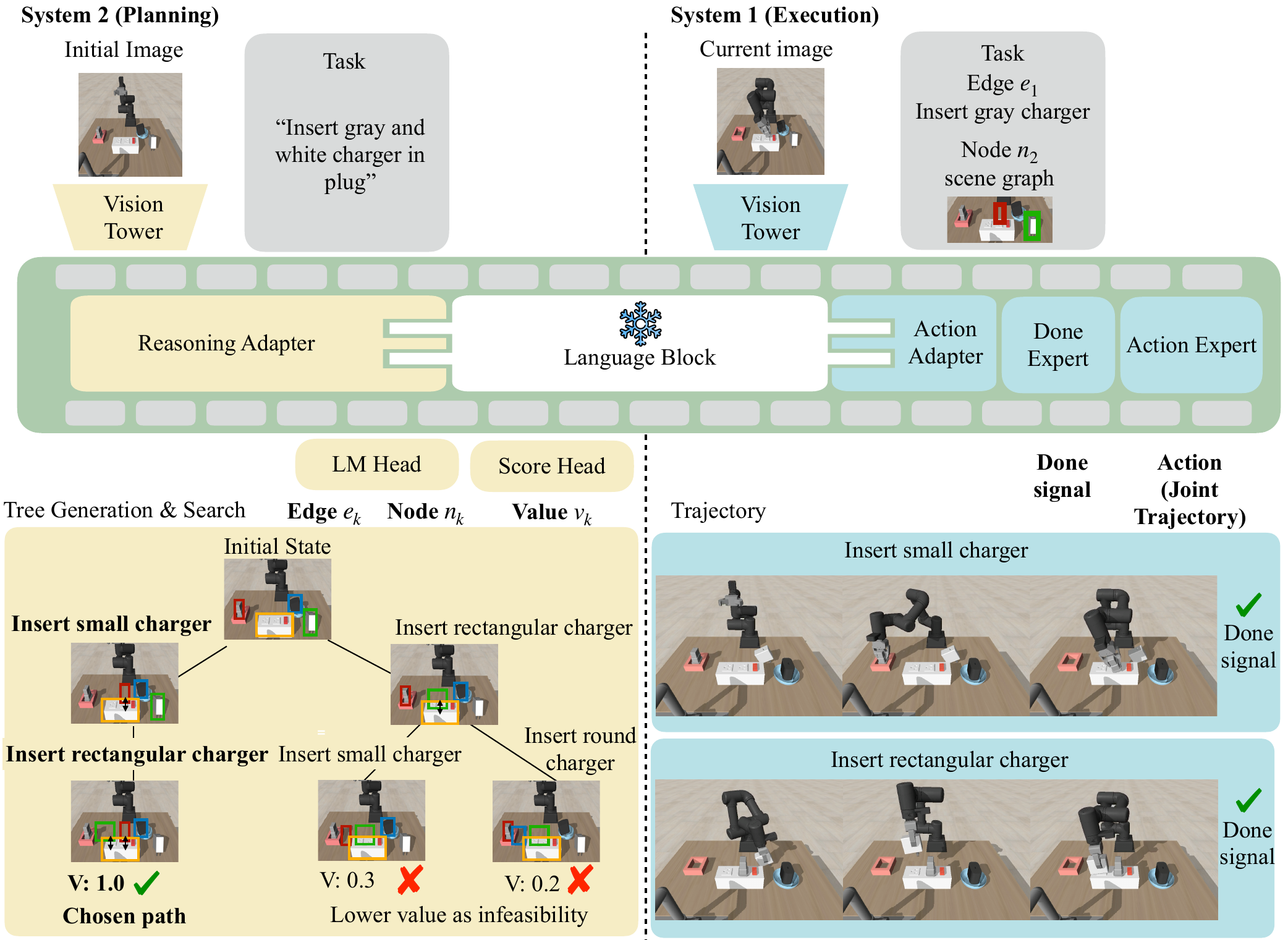}
    \caption{Architecture and overall pipeline. System~2 plans via tree-of-thoughts, predicting success values to select the highest-value path (green). System~1 executes the chosen subgoal as action chunks with a done signal. Both share a vision–language backbone with adapters.}
    \label{fig:arc}
\end{figure*}
In this section, we introduce \methodbold, a hierarchical \textbf{V}ision--\textbf{L}anguage--\textbf{A}ction framework that integrates negative experience into the \emph{reasoning} stage.
Our method is formulated within a hierarchical reinforcement learning~\cite{sutton1999between} framework consisting of two coupled systems.
\textbf{System~2 (Reasoning and Planning)} conducts tree search: proposes candidate reasoning paths, estimates their success probabilities, and selects the most feasible plan. \textbf{System~1 (Action Modeling)} then executes the next subgoal from the chosen plan with low-level actions.

\subsection{Overall Pipeline}
At inference time, the proposed framework follows a plan–execute procedure, illustrated in Figure~\ref{fig:arc}. 
Before acting, System~2 conducts a \emph{model-based tree search} over candidate options (edges) and their predicted successor states (nodes). Each candidate step is scored by how likely it is to succeed given the current scene and instruction, and the agent selects the highest-scoring overall path through the tree. 
System 1 then executes the current edge on that path with low-level actions until a termination condition indicates that it is complete. 
If execution matches the predicted successor state, the agent advances to the next edge on the chosen path. To meet real-time constraints, the tree is constructed before the execution starts in the initial state.

\subsection{Architecture}
Because System~2 performs \emph{model-based} planning over the abstract graph, it needs (a) a proposal mechanism for candidate edges, (b) a transition model to predict successor nodes, and (c) value heads to score lookahead, including failure awareness. Concretely, System~2 couples (i) an option generator $\hat P_\theta(e \mid n,z,\ell,s_0)$, (ii) a learned world model $\hat{P}_\theta(n'\mid e,n,z,\ell,s_0)$
and a value predictor $V(n) \coloneqq V_\theta(n|z,\ell,s_0)$, and (iii) a failure-aware value estimator $V(n) \coloneqq V_\theta(n|z,\ell,s_0)$ , yielding the high-level option policy. 
At the low level (System~1), the intra-option execution is parametrized by the 
intra-option policy $ \pi_e(a\mid s,\ell) \coloneqq \pi_\theta(a\mid s,\ell,e,n') $ 
and termination rule $\beta_e(s)  \coloneqq \beta_\theta(s,\ell,e,n')$.


We build the model upon \(\pi_0\)~\cite{black2024pi_0} as a multimodal backbone that encodes \((s,\ell,a)\) into a unified token sequence. Because \(\pi_0\) targets action generation rather than text, we merge its shared layers with a PaliGemma~\cite{beyer2024paligemma} language trunk, yielding strong text generation for tree-search steps while preserving \(\pi_0\)’s action priors~\cite{wortsman2022model}.
Both systems add LoRA~\cite{hu2022lora} adapters to this shared backbone. 
Specifically, we linearly interpolate the shared layers, 
$\theta^{\mathrm{merge}}_{g}(\lambda_m)=\lambda_m\,\theta^{\pi_{0}}_{g}+(1-\lambda_m)\,\theta^{\mathrm{PG}}_{g}$, 
\emph{where} $\theta^{\pi_{0}}_{g}$ and $\theta^{\mathrm{PG}}_{g}$ denote the parameters of the shared backbone from the pretrained $\pi_0$ and the PaliGemma, respectively, and $\lambda_m\in[0,1]$ controls the interpolation weight. 
The non-overlapping heads are kept separate.
\textbf{System~1} feeds adapted features to (i) an \emph{action expert} for low-level motor chunks and (ii) a \emph{done expert} that detects subgoal completion. 
System 1 has a shared attention layer between the language block and experts, where the done expert and the action expert do not attend to each other. 
\textbf{System~2} uses (i) a language-modeling head to autoregress node/edge text and (ii) a scalar value head to predict leaf value based on the last layer feature of the language backbone. 
Each system is trained independently with seperate dataset $\mathcal{D}_{sys1}$ and $\mathcal{D}_{sys2}$. 
This shares perception–language representations while keeping reasoning and control decoupled. 

\subsection{System 2}
System 2 is in charge of high-level tree-based planning, which is formulated as a meta-controller in HRL. 
Concretely, we instantiate it as a tree–search planner operating on a lookahead tree 
$\mathcal{T}=(\mathcal{N},\mathcal{E})$, where each node $n\in\mathcal{N}$ is a 2D scene graph and each edge $e\in\mathcal{E}$ denotes a verifiable subgoal that induces a transition $n_k \xrightarrow{\,e_k\,} n_{k+1}$. 
Thus, the abstract option–selection in HRL is realized by 
(1) \emph{successor–state (node) prediction}, 
(2) \emph{subgoal (edge) candidate proposal} labeling the transition $n_k\!\to\!n_{k+1}$, 
(3) \emph{leaf evaluation} with a value function $V_\theta$, and 
(4) \emph{tree search} to pick a feasible high-level plan. 
Given the initial scene $s_0$ and an instruction $\ell$, System~2 outputs a reasoning path 
$\tau^{\text{sel}}=(n_0,e_0,\dots,n_K)$ to hand over to System~1 for execution.


\subsubsection{Node and Edge Generation}
Edge proposals and next node transitions are generated autoregressively by the Language Modeling (LM) head, $\hat P_\theta$. The model first samples a candidate node string $n_{k+1}$, representing the next scene graph state, and then an edge string $e_{k+1}$, describing the subgoal transition leading there. This process follows the standard language modeling objective, factorizing the probability into token-level likelihoods: node $\hat P_\theta(n_{k+1}\mid z_{0:k},\ell,s_0) = \prod_{j=1}^{N_c} \hat P_\theta(w_j \mid w_{<j}, z_{0:k}, \ell,s_0)$ and edge $\hat P_\theta(e_{k+1}\mid n_{k+1}, z_{0:k}, \ell,s_0) = \prod_{j=1}^{N_e} \hat P_\theta(w_j \mid w_{<j}, n_{k+1}, z_{0:k}, \ell,s_0)$, 
where $N_c$ and $N_e$ denote the number of tokens for the node and edge string. 
In practice, we apply token-level beam search~\cite{freitag2017beam} to obtain multiple diverse candidates, and filter them with syntactic templates and scene-graph consistency checks (e.g., whether referenced objects exist in the current graph).
This ensures that generated subgoals remain both linguistically valid and grounded in the task context.  

We train with the standard LM loss over high-level tree-search step traces $(n_k,e_k,n_{k+1},z_{0:k-1},\ell,s_0) \sim \mathcal{D}_{\text{sys2}}$:
\begin{multline*}
\mathcal{L}_{\text{LM}}
= -\mathbb{E}_{\mathcal{D}_{\text{sys2}}}
\Big[
\log \hat P_\theta(n_k\mid z_{0:k-1},\ell,s_0) \\
+ \log \hat P_\theta(e_k\mid n_k,z_{0:k-1},\ell,s_0)
\Big].
\end{multline*}

\subsubsection{Node Evaluation}

We aim to estimate, at each search node, a calibrated first-exit success probability that steers expansion and pruning during tree search.
Each edge $e_k$ yields the successor node $n_{k+1}=\phi(s_{t_{k+1}})$; we label the transition success if $s_{t_{k+1}}\in\mathcal G$, failure if $s_{t_{k+1}}\in\mathcal F$, and continue otherwise, updating the context to $(n_{k+1},z_{0:k})$.



To define the training target, we use first-exit bootstrapping. For each transition $(n_k,e_k,n_{k+1},r_{k+1})$ we set
\begin{equation*}
y_k= \begin{cases} 1, & \mathclap{\text{terminal in }\mathcal G\ \big(\,e_k=\langle\texttt{done}\rangle\,\big)}\\ 
0, & \mathclap{\text{terminal in }\mathcal F\ \text{at step }k{+}1}\\ 
\gamma\,V_{\theta'}(n_{k+1}\mid z_{0:k},\ell,s_0), & \text{otherwise,} \end{cases} \end{equation*}
Here $\gamma\in(0,1)$ is the discount factor and $\theta'$ is a target network updated by an exponential moving average (EMA).

Based on Proposition \ref{sec:proposition}, the estimated value becomes the success probability of the node, 
\[
V_\theta(n_k\mid z_{0:k-1},\ell,s_0)\;=\;\Pr\!\big(\tau_{\mathcal G}<\tau_{\mathcal F}\,\big|\,n_k,z_{0:k-1},\ell,s_0\big)
\]
, where $\tau$ is the hitting time of the success and failure set. 

To handle offline data with numerous failures, we leverage asymmetric expectile loss (cf.\ IQL~\cite{kostrikov2022offline}) that discourages overestimation:
\[
    \mathcal{L}_{\text{val}}
    = \mathbb{E}_{\mathcal{D}_{\text{sys2}}} \Bigl[  L_2^{\tau_e}\!\left( y_k - V_\theta(n_k\mid z_{0:k-1},\ell,s_0)\right) \Bigr]
\]
with $ L_2^{\tau_e}(u) = \bigl|\tau_e - \mathbbm{1}(u<0)\bigr|\,u^2$ and $\tau_e=0.7$, which yields a conservative yet effective value from static datasets.

\subsubsection{Tree Search Algorithm}
We run a batched MCTS-style planner over abstract nodes (world states) and edges (subgoals). At each iteration, the planner selects promising frontier nodes by their running mean value $Q$, proposes and scores candidate edges, evaluates the resulting children with a learned state-value $V$, and backs up values along the path. After $M$ iterations, we return the plan traced from the highest-valued leaf, and execute the next edge of that plan.

Our search strategy is based on Monte Carlo Tree Search (MCTS) but replaces random rollouts with a learned state-value function $V(n)$. The algorithm builds a lookahead tree where nodes are world states and edges are subgoals, and it is accelerated via batched GPU computation. This procedure repeats for a fixed number of iterations $M$. Finally, we select the solution as the path leading to the leaf with the highest value, $\tau^{\text{sel}} = (n_0, e_0, \dots, n_K)$.

\begin{algorithm}[t]
\caption{\methodbold: Batched Tree Search with Learned $V$}
\label{alg:batch_mcts_short}
\SetKwInOut{Input}{Input}\SetKwInOut{Output}{Output}
\Input{Root $n_0$, batch $B$, proposals $k$, mix $\alpha$, steps $T$}
\Output{Plan (edge sequence) from the best frontier node}
Initialize tree $\mathcal{T}\!=\!\{n_0\}$; for all $n$: $N(n){=}0,\,W(n){=}0,\,Q(n){=}0$\;
\For{$m{=}1$ \KwTo $M$}{
  $\mathcal{F}\!\leftarrow$ frontier leaves; $\mathcal{S}\!\leftarrow$ Top-$B$ of $\mathcal{F}$ by $Q$\;
  \ForEach{$n_p\!\in\!\mathcal{S}$}{
    Propose $k$ edges $\{e_i\}$ (beam search decoding~\cite{freitag2017beam}); compute $S(n_p,e_i){=}\alpha Q(n_p)+(1{-}\alpha)\hat P(e_i\!\mid\!n_p)$; keep tops\;
  }
  For kept $(n_p,e)$: create child $n_c$ (greedy); predict $V(n_c)$ \;
  For path nodes $n_a$ to $n_c$: $W(n_a){+}{=}\!V(n_c)$, $N(n_a){+}{=}1$, $Q(n_a){=}\!W(n_a)/N(n_a)$\;
}
\Return plan traced from $\arg\max_{n\in\mathcal{F}} Q(n)$
\end{algorithm}

\subsection{System 1}
System~1 focuses on \emph{action execution}: in our HRL formulation, it implements both the intra-option policy and the termination rule, learned jointly by a single network.
It receives as input the selected high-level plan 
$\tau^{\text{sel}}=(n_0,e_0,\dots,n_K)$ from System~2,
and is responsible for grounding each symbolic edge into continuous control. 
Conditioned on $(s_t,\ell,e_k,n_{k+1})$, System~1 executes an option via a unified architecture producing the policy $\pi_\theta(\cdot\mid s_t,\ell,e_k,n_{k+1})$ and termination $\beta_\theta(s_t,\ell,e_k,n_{k+1})$.
System~1 is trained only on successful demonstrations. Both the action and termination experts
read from the shared backbone, without cross-attention or feature exchange.

\subsubsection{Training}
The intra-option policy (action expert) $\pi_\theta$ is parameterized by a flow–matching model. We denote the low–level action chunk at control time $t$ as
\(\mathbf{A}_t  = \{a_t, \cdots, a_{t+H}\}\in \mathbb{R}^{H\times d_a}\). 
Let $\tau\in[0,1]$ denote the interpolation time, and $\epsilon$ a Gaussian noise variable.  
The noisy action is $\mathbf{A}_t^\tau \;=\; \tau\,\mathbf{A}_t + (1-\tau)\epsilon$.
The policy $\pi_\theta$ predicts the flow, 
$\dot{\mathbf A}_\theta(\mathbf{A}_t^\tau,s_t,\ell,e_k,n_{k+1})$,
and is trained to match the oracle velocity $\mathbf u(\mathbf{A}_t^\tau\mid \mathbf A_t)$:
\[
\mathcal{L}_{\text{act}}
=\mathbb{E}_{\mathcal{D}_{sys1}}
\Big\|
\dot{\mathbf A}_\theta(\mathbf{A}_t^\tau,s_t,\ell,e_k,n_{k+1})
-\mathbf{u}(\mathbf{A}_t^\tau\mid \mathbf A_t)
\Big\|_2^2.
\]

The done expert $\beta_\theta$ predicts the probability of end–of–edge,
$p^{\text{done}}_\theta=\beta_\theta(s_t,\ell,e_k,n_{k+1})$,
trained with focal loss 
\[
\mathcal{L}_{\text{done}}
=\mathbb{E}_{\mathcal{D}_{sys1}} \bigl[-\alpha_d\,(1-p^{\text{done}}_\theta)^{\gamma_d}\log p^{\text{done}}_\theta \bigr].
\]
The joint system~1 objective is
$
\mathcal{L}_{\text{S1}}
=\lambda_{\text{act}}\mathcal{L}_{\text{act}}
+\lambda_{\text{done}}\mathcal{L}_{\text{done}}
$, where $\lambda_{\text{act}}$ and $\lambda_{\text{done}}$ are weight hyperparameters. 


\subsubsection{Inference}
Given the active pair $(e_k,n_{k+1})$ from $\tau^{\text{sel}}$, we initialize
$\mathbf{A}_t^{0}\!\sim\!\mathcal{N}(0,\sigma_0^2 I)$ and integrate the learned flow along the interpolation path:
\[
\mathbf{A}_t^{\tau+\delta}
=\mathbf{A}_t^{\tau}
+\delta\,\dot{\mathbf A}_\theta(\mathbf{A}_t^\tau, s_t,\ell,e_k,n_{k+1})
\]
where $\tau\in\{0,\delta,\ldots,1-\delta\}$. 
After sweeping to $\tau=1$, we obtain the predicted action chunk
$\widehat{\mathbf{A}}_t=\mathbf{A}_t^{1}$.

In parallel, the termination head outputs the end-of-edge probability
$p^{\text{done}}_\theta=\beta_\theta(s_t,\ell,e_k,n_{k+1})$.
We advance to the next symbolic step when this probability exceeds a threshold
$\kappa_{\text{done}}$:
\begin{align*}
    k \leftarrow k + \mathbf{1}\!\left\{\,p^{\text{done}}_\theta \ge \kappa_{\text{done}}\,\right\}, \\
    (e_k,n_{k+1}) \leftarrow \text{next pair from } \tau^{\text{sel}}.
\end{align*}
If $k>K$, the option sequence terminates; otherwise, System~1 repeats the procedure for the updated pair.

%% file: sections/5.experiments.tex
\section{Experiment}
\begin{figure}
    \centering
    \includegraphics[width=0.99\linewidth]{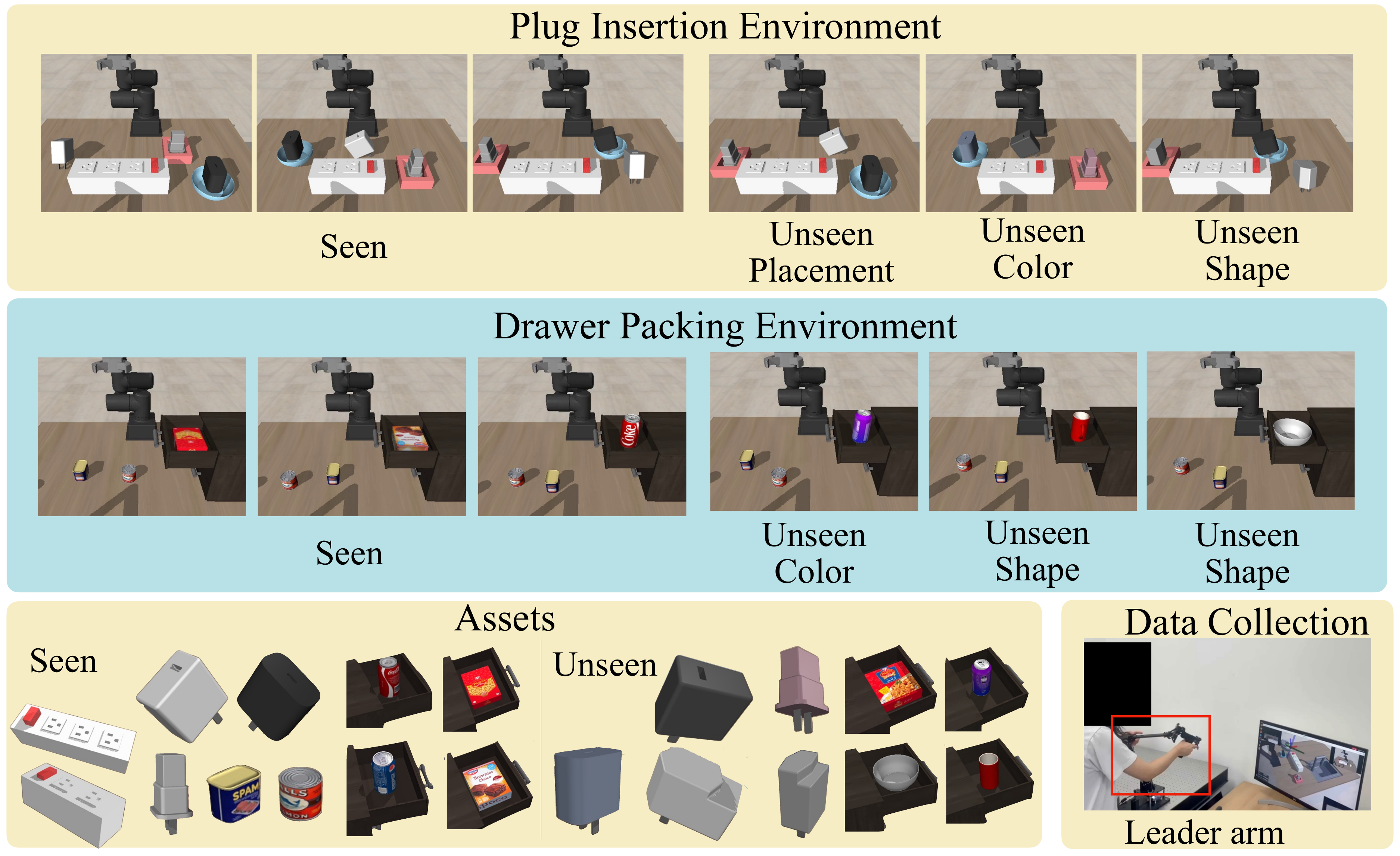}
    \caption{Plug insertion and drawer packing environment. We evaluate in both seen and unseen settings with corresponding assets, and collect demonstrations via a teleoperation setup.}
    \label{fig:env}
\end{figure}

\begin{table*}[]
    \centering
    \caption{Simulation Results in plug insertion and drawer packing environments, showing the success rate.}
    \resizebox{0.8\linewidth}{!}{%
    \begin{tabular}{lc|ccc|ccc}
    \toprule
    \multirow{2}{*}{Models} & \multirow{2}{*}{Failure Data} & \multicolumn{3}{c}{Plug Insertion} &  \multicolumn{3}{c}{Drawer Packing} \\ 
    & & Seen & Unseen & Average & Seen & Unseen & Average \\
    \midrule
    \textcolor{gray}{\textit{Unified Model}} & & & & \\ 
     OpenVLA-OFT~\cite{kim2025fine} & $\times$ & 0.244 & 0.044 & 0.144 & 0.637 & 0.283 & 0.485 \\
     GR00T N1.5~\cite{bjorck2025gr00t} & $\times$ & 0.422 & 0.244 & 0.333 & 0.704 & \underline{0.600} & 0.669 \\
    $\pi_0$~\cite{black2024pi_0} & $\times$ & 0.689 & 0.267 & 0.477 & 0.735 & 0.575 & 0.675\\
    $\pi_0$+ Reward Cond.~\cite{yuan2023rewarddirected} & $\checkmark$ & 0.489 & 0.111 & 0.300 & 0.550 & 0.425 & 0.508 \\
    \midrule
    \textcolor{gray}{\textit{SOTA VLM as System2}} & & & & \\ 
     GPT-4o~\cite{hurst2024gpt} & $\times$ & 0.733 & 0.311 & \underline{0.522} & 0.650 & 0.475 & 0.591 \\
     GPT-4o~\cite{hurst2024gpt} & $\checkmark$ & 0.711 & \underline{0.333} & 0.488 &\underline{ 0.738} & 0.575 & \underline{0.683} \\
     Gemini-2.5-Flash~\cite{comanici2025gemini} & $\times$ & \underline{0.756} & 0.200 & \underline{0.522} & 0.637 & 0.450 & 0.575  \\
     Gemini-2.5-Flash~\cite{comanici2025gemini} & $\checkmark$ & 0.711 & 0.289 & 0.500 & 0.713 & 0.525 & 0.650 \\
    \midrule
    \textcolor{gray}{\textit{Role of failure data in Our System2}} & & & & \\ 
    \method-Chain & $\times$ & 0.733 & 0.244 & 0.488 & 0.700 & 0.450 & 0.616  \\
    \method-Tree & $\times$ & 0.711 & 0.289 & 0.500 & 0.732 & 0.525 & 0.663  \\
    \rowcolor{blue!10}   
    \method-Full (Ours) & $\checkmark$ & \textbf{0.800} & \textbf{0.422} & \textbf{0.611} & \textbf{0.800} & \textbf{0.675} & \textbf{0.752} \\
    \bottomrule
    \end{tabular}%
    }
    \label{tab:results}
\end{table*}

In this section, we present a series of experiments that validate the effectiveness and generality of our framework, with an emphasis on how failure data biases plan selection toward higher-probability trajectories. 
Our evaluation is organized along four complementary parts. 
First, in \textbf{Simulation Environments and Data Collection} (Section~\ref{sec:settings}), we introduce teleoperated tasks, mixed-quality demonstrations, and seen/unseen splits; and in \textbf{Simulation Results and Analysis} (Section~\ref{sec:teleop}), we benchmark against unified VLA and VLM-as-planner baselines to quantify the effect of failure-aware reasoning. 
Second, in \textbf{Test-Time Scalability} (Section~\ref{sec:scalability}), we hold model parameters fixed and vary the planner’s per-step expansion width \(K\) to assess robustness gains from additional inference-time reasoning. 
Third, in \textbf{Bootstrapping Pretrained Models} (Section~\ref{sec:boot}), we reuse rollouts from a strong pretrained policy to retrain with our failure-aware objective and measure transfer. 
Fourth, in \textbf{Real-World Deployment} (Section~\ref{sec:realworld}), we execute the selected high-level chain on a physical robot. 
Across all parts, we report success rate as the primary metric and compare against imitation- and VLM-based baselines that do not explicitly utilize failure data.


\subsection{Simulation Environments and Data Collection}\label{sec:settings}
To the best of our knowledge, there is no widely adopted benchmark that makes failure-aware evaluation of spatial reasoning straightforward; most existing resources only provide success demonstrations. 
We therefore introduce a new environment suite, where plug insertion and drawer packing are illustrative instances, with controlled geometry, contact, and placement variations under explicit seen/unseen splits.
\subsubsection{Environment Setup}
The environment is built upon a MuJoCo~\cite{6386109} simulator, as illustrated in Figure~\ref{fig:env}.  
Plug insertion (seen: 9 configs) combines three table placements with 2- or 3-socket strips and orientation variants; \emph{unseen} splits introduce novel charger placement, unseen strip color, and unseen charger shape. The strip collision margin is slightly relaxed to reduce teleop burden. 
\emph{Drawer packing} (seen: 12 configs) pairs four table placements with two distractor types (a box and a can) placed inside the drawer; \emph{unseen} splits (4 configs) alter distractor color and type. Unlike plug insertion, which is mostly constrained by insertion order, drawer packing admits multiple feasible behaviors (push the distractor aside, pick and remove it, or leave it), making contact outcomes highly \emph{stochastic}. Small variations in contact geometry, stick–slip, and drawer clearance lead to large success variance even under near-identical initial states.
Control runs at 20\, Hz with joint-position actions.

\subsubsection{Data collection}
We gather human teleoperation trajectories via a leader–follower setup (Fig.~\ref{fig:env}). 
For \emph{plug insertion}, we collect 450 demonstrations over predefined insertion orders (six paths for 3-socket, two for 2-socket), balanced per path. Each trajectory is labeled success/failure; outcomes are naturally stochastic but show clear mode structure: PlugStrip-3 Fail dominates (38.9\%, 175/450), followed by PlugStrip-2 Success (30.4\%, 137), PlugStrip-3 Success (21.1\%, 95), and PlugStrip-2 Fail (9.6\%, 43), yielding a multi-modal dataset whose success rate varies with insertion order. 
For \emph{drawer packing}, we collect 240 demonstrations in total—120 per distractor type (box/can)—covering three strategies (\textsc{Pick}, \textsc{Push}, \textsc{Leave}). Compared to plug insertion, outcomes are strongly stochastic: success probabilities depend sharply on both strategy and distractor identity (e.g., \textsc{Leave} is often brittle when the can closely occludes the handle, whereas \textsc{Pick}/\textsc{Push} are more reliable but still contact-sensitive). This task, therefore, provides diverse, multi-modal successes and failures arising from real-world manipulation affordances rather than a single fixed plan. The detailed statistics of the dataset are shown in the Appendix. 


\subsubsection{Baselines}

To assess our two-system framework, we compare against three groups: (i) unified VLA models, (ii) VLM-as-planner baselines that replace System2 while keeping System1 fixed, and (iii) our System~2 variants (Chain/Tree/Full) to isolate branching and failure-conditioned value. Since our environment operates at 20Hz in a joint-position action representation, we select baselines that can be executed under this control frequency and action format.

\begin{figure*}
    \centering
    \includegraphics[width=0.99\linewidth]{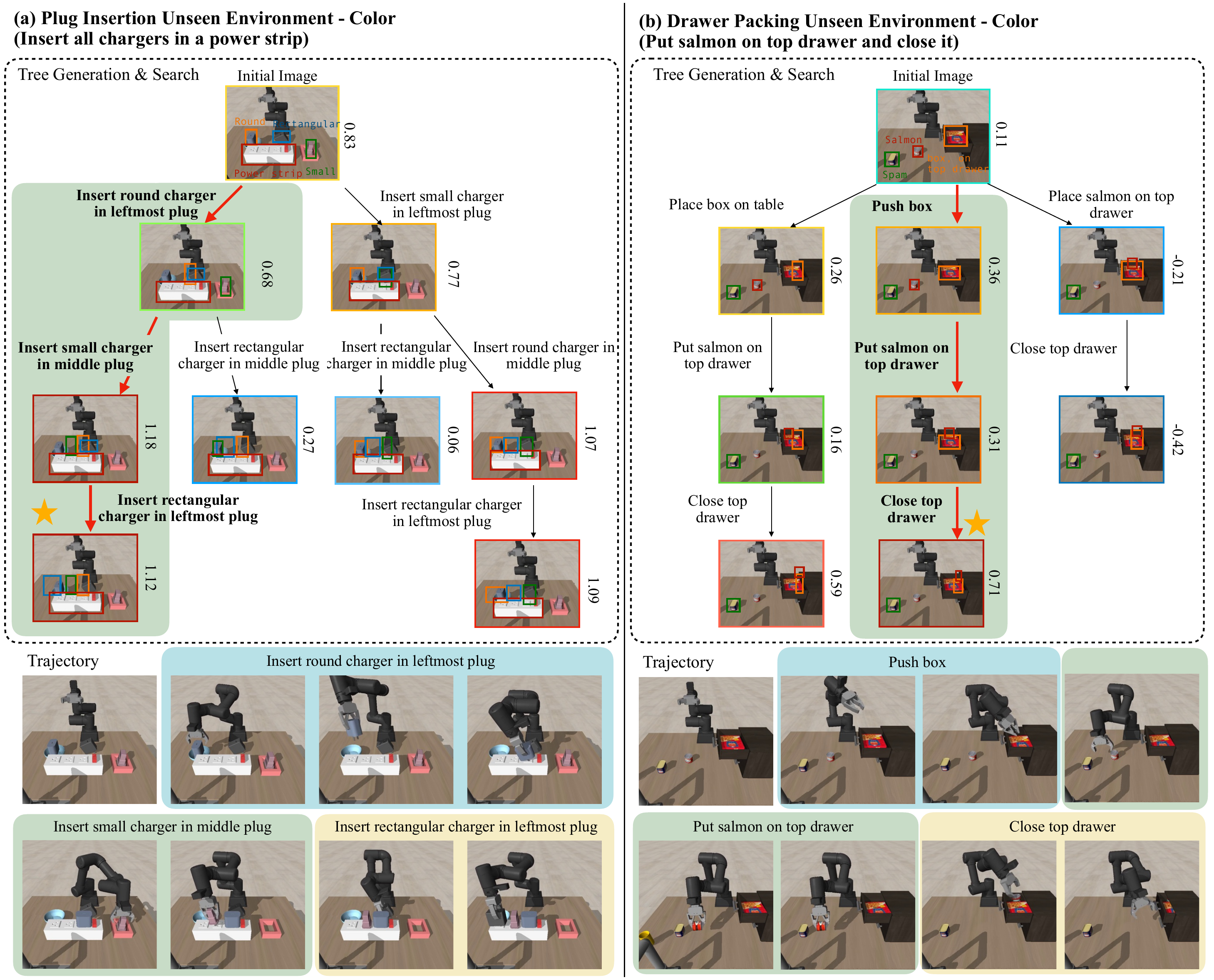}
    \caption{Qualitative results. System2 builds a failure-aware planning tree with feasibility scores, and System1 executes the selected subgoals, producing successful trajectories.}
    \label{fig:result}
\end{figure*}

\begin{itemize}
\item \textbf{Unified VLA models.} OpenVLA-OFT~\cite{kim2025fine}, GR00T~N1.5~\cite{bjorck2025gr00t}, and $\pi_{0}$~\cite{black2024pi_0}, plus a reward-conditioned $\pi_{0}$ variant using failure data~\cite{yuan2023rewarddirected}.
\item \textbf{VLM-as-System~2.} Replace our planner with SOTA VLMs (e.g., GPT-4o~\cite{hurst2024gpt}, Gemini-2.5-Flash~\cite{comanici2025gemini}) using few-shot prompting to propose subgoals/next scene graphs, with and without failure examples; System~1 (our action executor) is fixed.
\item \textbf{Our System~2 variants.} \method-Chain (no branching without failure data), \method-Tree (tree search without failure data, scoring with confidence), and \method-Full (tree search + failure-conditioned value).
\end{itemize}

\subsection{Simulation Results and Analysis}\label{sec:teleop}
We evaluate all models on success rate across both seen and unseen configurations of the plug insertion and drawer packing tasks. The results, summarized in Table~\ref{tab:results}, demonstrate that our proposed model, \method, consistently and significantly outperforms all baselines. Our analysis delves into the performance comparison against state-of-the-art models and investigates the impact of failure data by comparing variants of our own system that share the same underlying architecture.

\subsubsection{Comparison with Unified VLA Models}
Unified VLA models, which operate as end-to-end reactive policies, demonstrate a critical weakness in generalization. While some models like $\pi_0$~\cite{black2024pi_0} achieve reasonable performance on seen configurations (0.689 in plug insertion), all unified VLAs experience a severe performance degradation when faced with novel scenarios. For instance, $\pi_0$'s success rate plummets from 0.689 to 0.267 in the unseen plug insertion task. Similarly, OpenVLA-OFT~\cite{kim2025fine} drops from 0.244 to a mere 0.044, and GR00T N1.5~\cite{bjorck2025gr00t} falls from 0.422 to 0.244. This consistent trend highlights their difficulty when learning long-horizon tasks from multi-modal demonstration data: they struggle to discern which of the many possible paths is most feasible. Lacking a mechanism for deliberation, they cannot effectively weigh alternative strategies, a challenge that our two-system architecture is designed to overcome.
 \begin{figure*}
    \centering
    \includegraphics[width=0.9\linewidth]{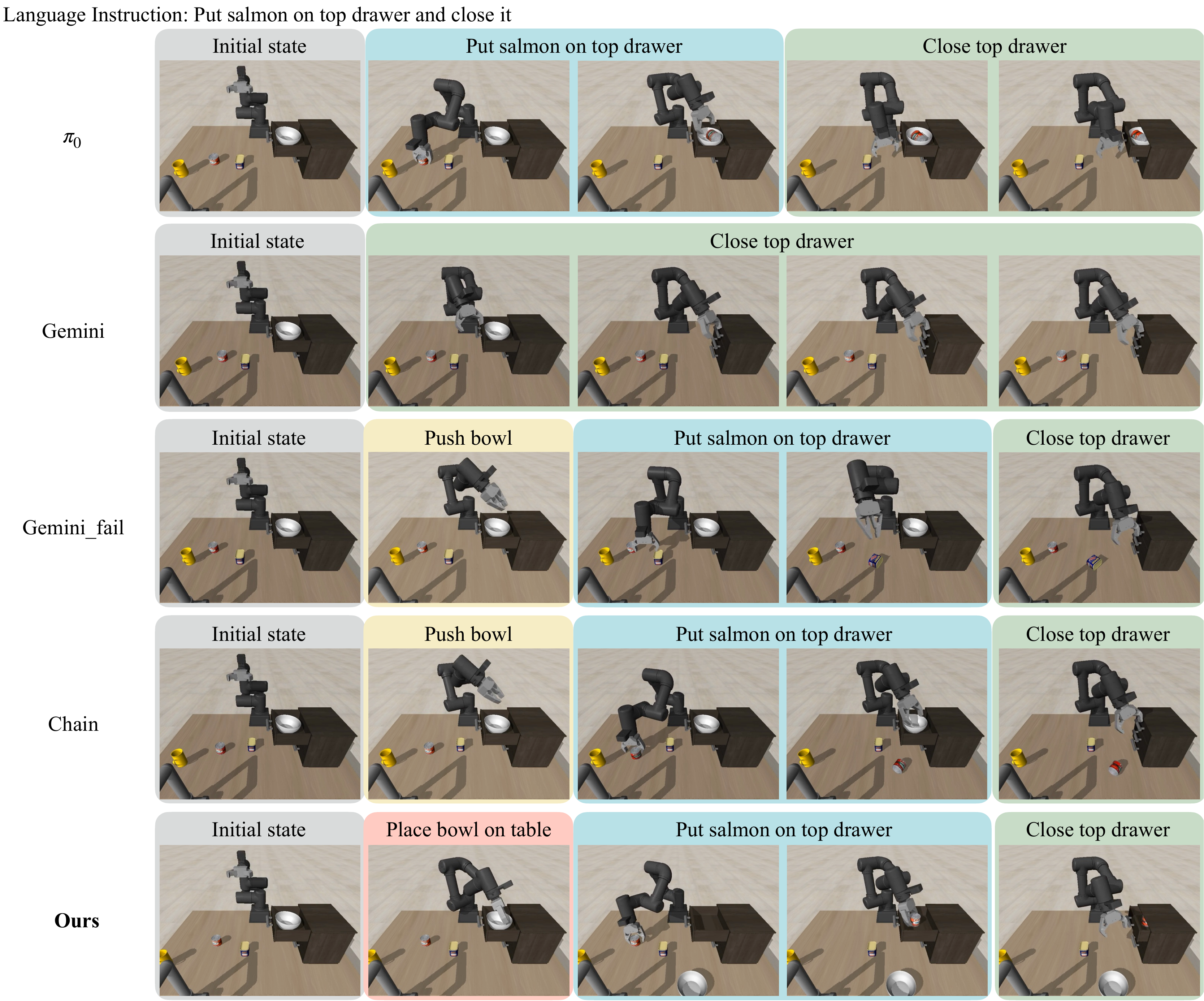}
    \caption{Trajectory Comparison with baselines. The figure shows the trajectory generated by $\pi_0$~\cite{black2024pi_0}, VLM as system 2, i.e., Gemini-2.5-flash~\cite{comanici2025gemini} (Gemini), and with failure examples (Gemini fail), \method-Chain (chain), and \method-Full (Ours) in an unseen drawer packing environment.}
    \label{fig:res_comparison}
\end{figure*}
\subsubsection{Comparison with VLM-as-Planner Baselines}
Using VLMs as the planner separates deliberation from control and yields higher average performance than unified VLAs. Even so, \method-Full remains stronger, especially under novelty. On drawer packing, \method-Full achieves 0.752, a relative gain of 10.1\% over the best VLM baseline (GPT-4o with failure examples~\cite{hurst2024gpt}, 0.683), and 0.675 on the unseen split compared with 0.575. On plug insertion, \method-Full attains 0.611 versus 0.522 for GPT-4o/Gemini-2.5-Flash~\cite{comanici2025gemini}, and 0.422 on the unseen split versus 0.333, corresponding to 17.1\% and 26.7\% relative improvements, respectively. Adding failure examples to prompts improves VLMs (for example, GPT-4o on unseen drawer increases from 0.475 to 0.575), but their grounding in task dynamics remains limited.
The VLM approach relies on in-context learning to adapt its general knowledge from a few textual examples, which provides strong reasoning but is not deeply grounded in the task's specific physical realities. This can lead to conceptually sound plans that overlook subtle failure modes, a vulnerability particularly exposed in novel settings. 
In contrast, our method's components are trained directly on demonstration data, creating a more grounded signal for a tree search. This allows our planner to assess a plan's feasibility by exploring its predicted consequences, making it more robust in novel scenarios.

\subsubsection{Role of Failure Data and Tree Search}
Our component analysis reveals that both tree search and failure-conditioning are critical to our model's success. The \method-Chain model, a linear planner using only success data, struggles to generalize (0.244 unseen success). While leveraging tree search (\method-Tree) improves unseen performance by allowing the model to explore alternatives, which is especially valuable in the multi-modal drawer packing task (unseen success jumps from 0.450 to 0.525). 
However, the most significant improvement comes from training the value function with failure data, which provides a signal to judge a plan's feasibility. 
\method-Full boosts unseen success rates to 0.422 in plug insertion and 0.675 in drawer packing. This result decisively shows that a value function that internalizes the dynamics of failure provides a more reliable planning signal than the generalized confidence of system 2, making it key to robust performance in novel scenarios.

\subsubsection{Qualitative Results}
Figure~\ref{fig:result} shows qualitative results of our model's planning and execution capabilities on two distinct, complex tasks. In the plug-initiation task (a), it successfully determines the correct multi-step sequence for inserting various chargers into their corresponding slots. For the Drawer Packing task (b), it demonstrates strategic reasoning by first pushing an obstacle to make space before placing the target object. The successful roll-out trajectories for each task validate that our system can effectively execute these complex, long-horizon plans. 
Furthermore, Figure~\ref{fig:res_comparison} contrasts trajectories on drawer packing. Reactive baselines,$\pi_0$~\cite{black2024pi_0} and Gemini-2.5-Flash~\cite{comanici2025gemini} as System 2, ignore the obstructing bowl and attempt a direct placement, leading to collisions. An ablation using only success data without tree search (“chain”) merely tries to shove the obstacle aside. In contrast, our model plans a feasible, multi-step solution: first relocate the bowl to the table, then place the salmon in the drawer, demonstrating stronger strategic reasoning and execution.

\begin{figure}[!t]
    \centering
    \includegraphics[width=0.95\linewidth]{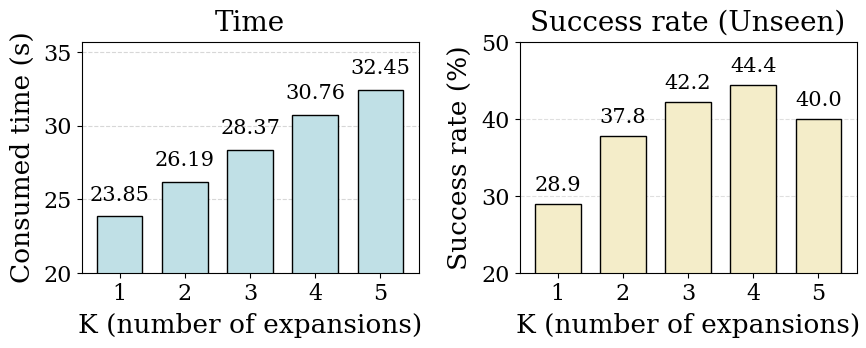}
    \caption{Tree Analysis for test time scaling across per-step expansion. In the unseen plug-insertion setting, increasing the search width \(K\) consistently improves success while increasing inference time, demonstrating a test-time scalability.}
    \label{fig:test_time_scale}
\end{figure}
\begin{table}[!t]
    \centering
    \caption{Results from Simpler Environment. Carrot. denotes carrot on plate, eggplant. denotes eggplant in basket, and spoon. for spoon on towel. }
    \begin{tabular}{c|cccc|c}
        \toprule
        Method & carrot. & eggplant. & cube. & spoon. & Average \\
        \midrule
        $\pi_0$ (Finetuned) & 0.236 & \textbf{0.791} & 0.139 & 0.388  & 0.389 \\
        \rowcolor{blue!10}  
        \method (Ours) & \textbf{0.375} & 0.708  & \textbf{0.145}  & \textbf{0.395} & \textbf{0.406} \\
        \bottomrule
    \end{tabular}
    \label{tab:simpler}
\end{table}

\subsection{Test-Time Scalability}\label{sec:scalability}

Humans naturally allocate more time to difficult problems, adjusting their reasoning effort to the complexity of the situation. In contrast, most VLA models commit to a single forward pass per observation, applying a fixed amount of computation regardless of task uncertainty. We hypothesize that allocating additional inference-time reasoning, by expanding and verifying multiple subgoal hypotheses, can improve robustness without retraining. This setting allows us to study whether compute adaptivity at deployment can serve as an additional axis of generalization.

To evaluate this, we vary the planner’s per-step expansion width \(K\), which controls how many candidate subgoal branches are explored and scored before acting (Figure~\ref{fig:test_time_scale}) in unseen configurations of plug-insertion environment. Increasing \(K\) expands the search tree and enables broader reasoning over plausible alternatives, while keeping model parameters fixed. As \(K\) increases, unseen success improves \(28.9\%\!\to\!37.8\%\!\to\!42.2\%\!\to\!44.4\%\!\to\!40.0\%\) for \(K{=}1,2,3,4,5\), with a near-linear latency growth of \(23.9\text{s}\!\to\!26.2\text{s}\!\to\!28.4\text{s}\!\to\!30.8\text{s}\!\to\!32.5\text{s}\). These results demonstrate that modest increases in test-time reasoning yield measurable improvements in success under distribution shift, reflecting a favorable accuracy–latency trade-off and diminishing returns beyond \(K{=}4\).


\subsection{Bootstrapping Pretrained Models}\label{sec:boot}

\begin{figure}[!t]
    \centering
    \includegraphics[width=0.99\linewidth]{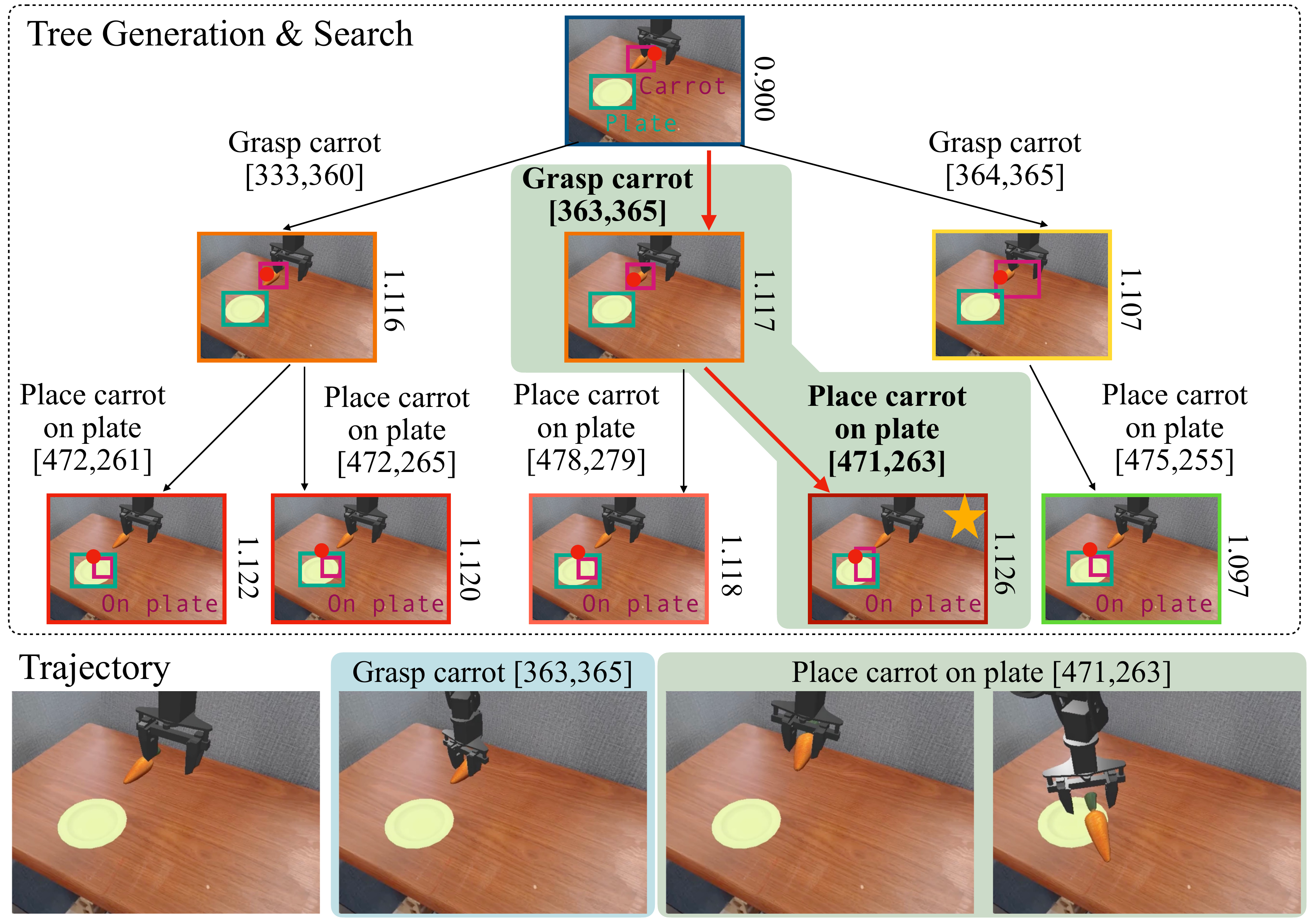}
    \caption{Generated tree and trajectories in Simpler environment.}
    \label{fig:simpler}
\end{figure}

In this experiment, we bootstrap a pretrained policy by rolling out a Bridge v2~\cite{walke2023bridgedata}–finetuned $\pi_0$ to collect 600 additional raw trajectories, then re-train with our failure-aware procedure and merge the adapted weights (PaliGemma backbone + finetuned $\pi_0$). We evaluated the proposed method in the Simpler environment~\cite{li2024evaluating} with the Widow-X robot in four different tasks: carrot. (\textit{put carrot on plate}), eggplant. (\textit{put eggplant in basket}), cube. (\textit{stack green block on yellow block}) and spoon (\textit{put spoon on towel}). In this environment, we define edge as a subgoal with a 2D end-effector position in the image, as there does not exist multiple available subgoals to succeed the tasks. To balance the dataset between four tasks, we collected more data on lower success rate tasks like \textit{stack cubes} or \textit{carrot on plate}.

As summarized in Table~\ref{tab:simpler}, our method yields a modest but consistent gain in average success rate on the \textit{Simpler} tasks (from 0.389 to  0.406), with clear improvements on \textit{carrot on plate} (0.236 to 0.375), \textit{stack cubes} (0.139 to 0.145), and \textit{spoon on towel} (0.388 to 0.395). The only regression appears on \textit{eggplant in basket} (0.791 to 0.708), where the baseline already exhibited a very high success rate, suggesting diminishing returns and occasional overfitting or value recalibration when bootstrapping in near-saturated regimes. Overall, the results indicate that rolling out a trained policy and re-training it with our failure-informed updates can improve robustness and transfer to previously weaker objects while preserving strengths elsewhere. 
The example of a generated tree and accompanying trajectory is illustrated in Figure~\ref{fig:simpler}.

\begin{figure}
  \centering
  \subfloat[Real-world workspace and setup.\label{fig:rl_setting_sub}]{
    \includegraphics[width=0.95\linewidth]{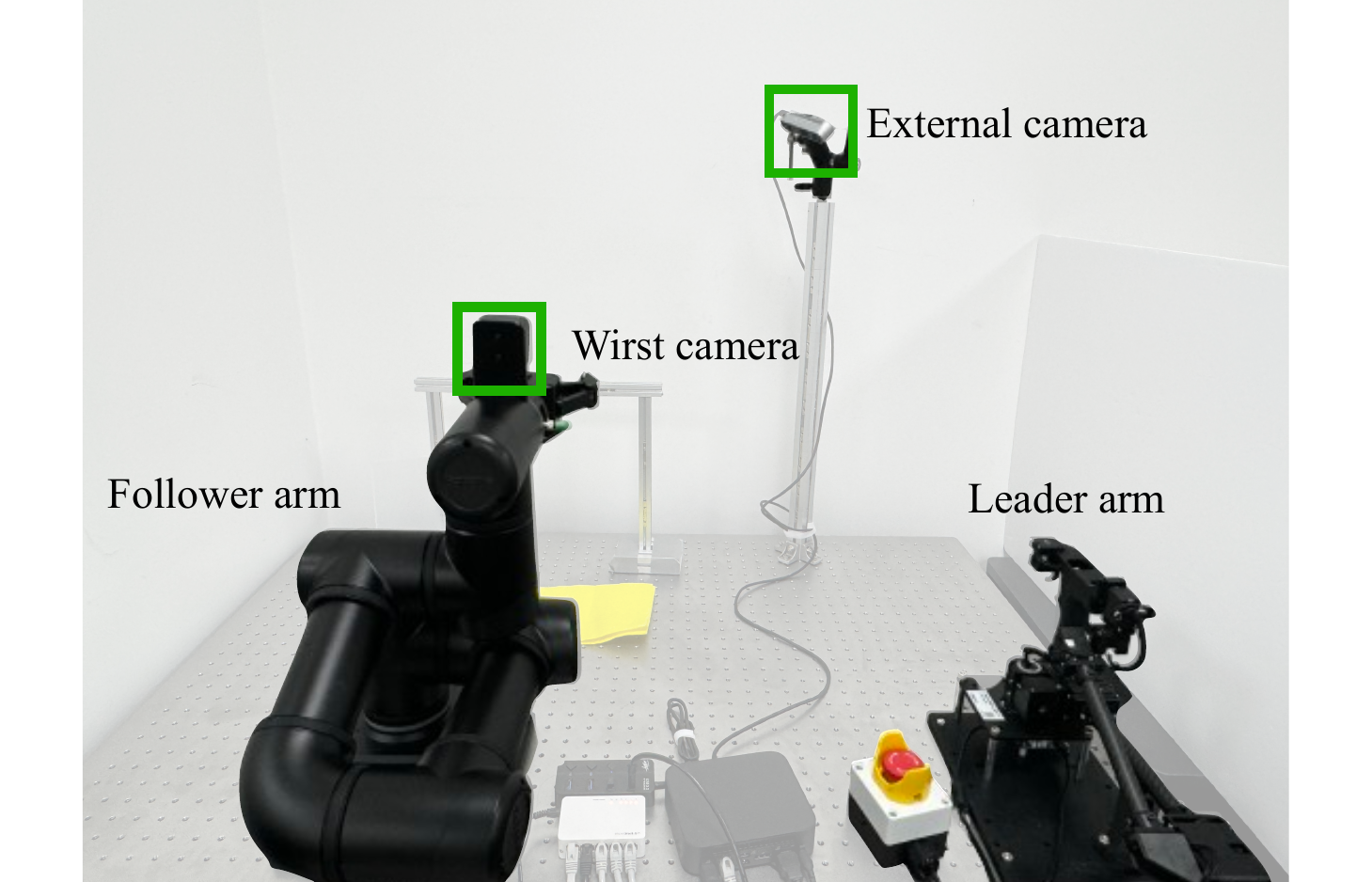}}\par
  \vspace{-0.5em}
  \subfloat[Tasks used in evaluation.\label{fig:rl_tasks_sub}]{
    \includegraphics[width=0.95\linewidth]{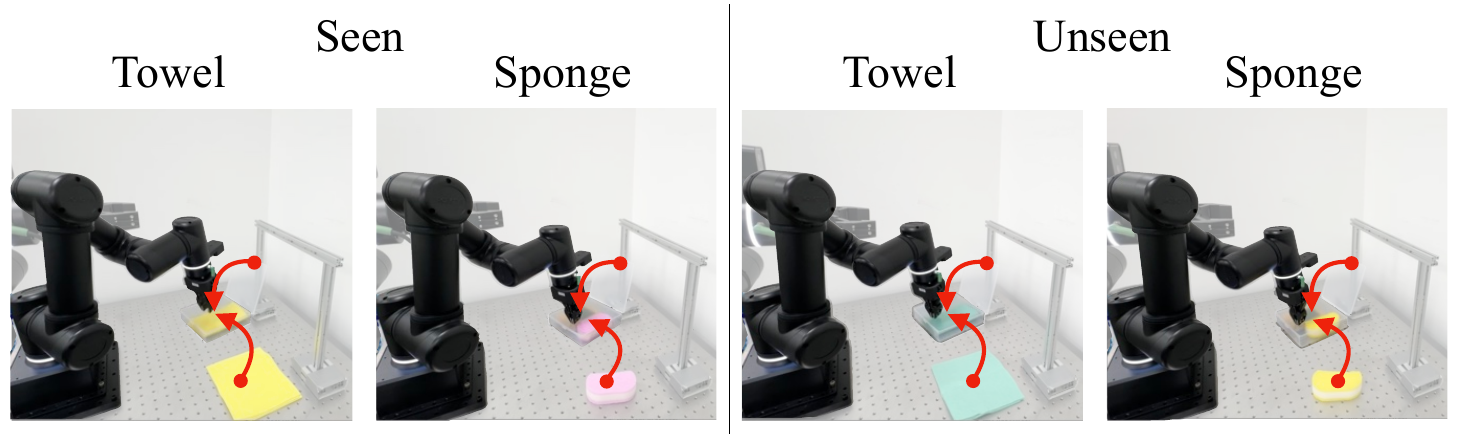}}
  \caption{Real-world environment and tasks.
\textbf{(a)} Setup with a 6-DoF arm (joint-position control, wrist + external cameras).
\textbf{(b)} Tasks: sponge and towel packing in the cabinet, with distractors and folding strategies.
Generalization is tested from \emph{seen} to \emph{unseen} color configurations.}
  \label{fig:rl_tasks}
\end{figure}
\subsection{Real-world Demonstration}\label{sec:realworld}

\begin{figure*}
    \centering
    \includegraphics[width=0.95\linewidth]{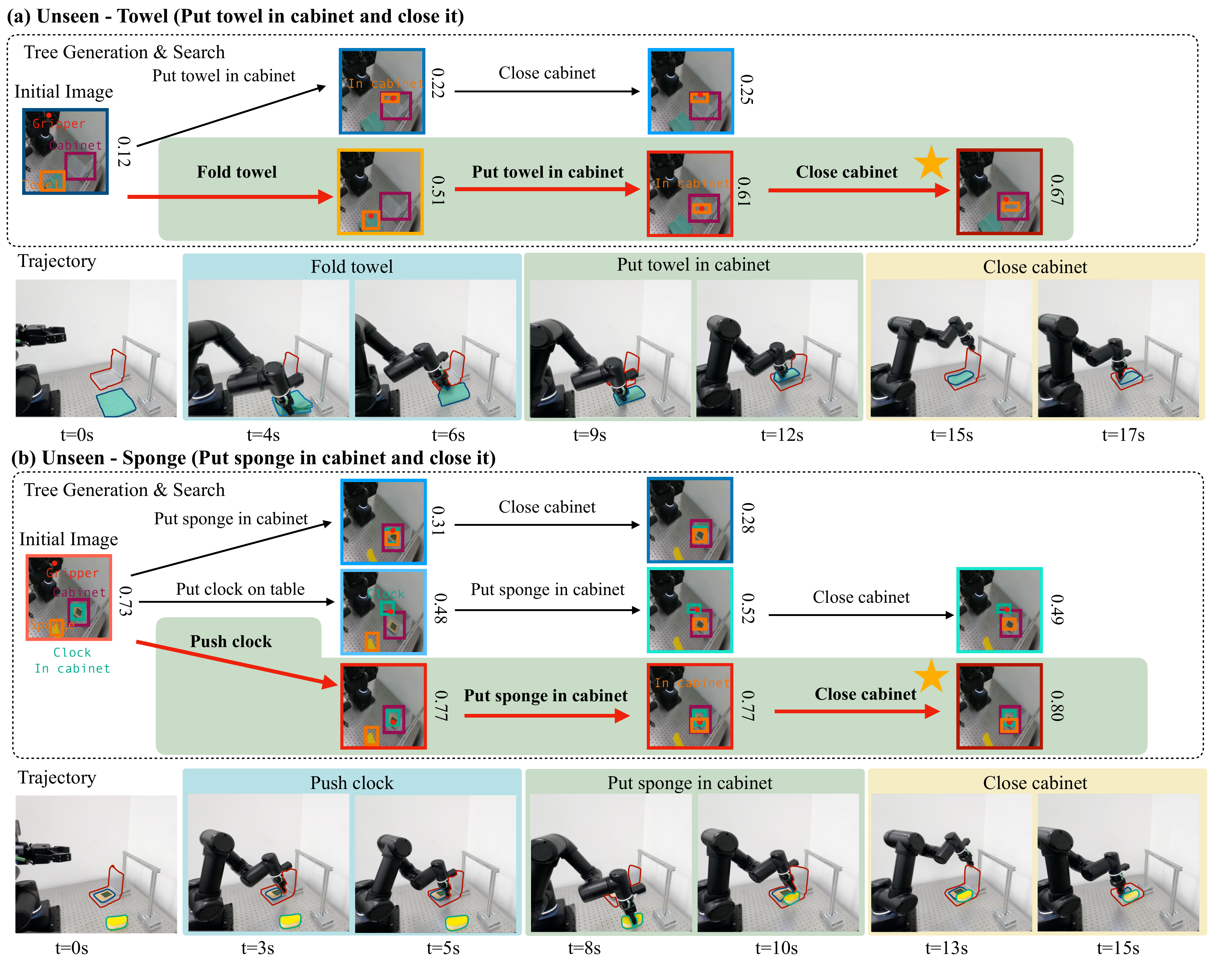}
    \caption{Real-world demonstrations with planning and execution.
\textbf{(a)} \emph{Towel–unseen}: the search tree selects \textit{fold} $\rightarrow$ \textit{put in cabinet} $\rightarrow$ \textit{close}, then System~1 executes the chosen path.
\textbf{(b)} \emph{Sponge–unseen} (clock distractor): the planner chooses \textit{push clock} $\rightarrow$ \textit{put sponge} $\rightarrow$ \textit{close} and the trajectory follows accordingly.}

    \label{fig:rl_demo}
\end{figure*}
\begin{table}[!t]
    \centering
    \caption{Real-world success rates on sponge and towel packing tasks. Our method (\method) outperforms the \(\pi_{0}\) baseline in both seen and unseen settings.}
    \begin{tabular}{c|cc|cc}
        \toprule
        \multirow{2}{*}{Method} & \multicolumn{2}{c|}{Seen} & \multicolumn{2}{c}{Unseen}  \\
        & sponge & towel & sponge & towel \\
        \midrule
        $\pi_0$~\cite{black2024pi_0} & 0.60 & 0.45 & 0.55 & 0.30 \\
        \rowcolor{blue!10}  
        \method (Ours) &  \textbf{0.75} & \textbf{0.55} & \textbf{0.65} & \textbf{0.55} \\
        \bottomrule
    \end{tabular}%
    \label{tab:rlworld}
\end{table}

\subsubsection{Environment Setting and Tasks}
Our experiments are conducted using a 6-DoF robotic arm controlled at a 20Hz frequency with joint position actions, receiving visual input from both a wrist-mounted and a fixed third-person camera. We introduce two multi-modal, long-horizon tasks that require placing an object into a container and subsequently closing its lid. The environment contains two different language instructions. The first task, sponge placement, involves placing a sponge inside a container already occupied by a clock, which admits three distinct solution strategies: pushing the clock aside, removing it from the cabinet entirely, or placing the sponge directly into the available free space. The second task, towel placement, involves placing a towel, which can be completed with or without an optional intermediate folding step. To evaluate generalization, we define \emph{seen} configurations using a pink sponge and a yellow towel, and test on \emph{unseen} configurations that introduce a domain shift via a yellow sponge and a green towel, as depicted in Figure~\ref{fig:rl_tasks}.
We collect 20 expert demonstrations for each of the five distinct strategies, for a total of 100 trajectories.

\subsubsection{Results}
We evaluate our method, \method, against the baseline on the real-world manipulation tasks, with quantitative results presented in Table \ref{tab:rlworld}. 
For seen objects, our model achieved success rates of 75\% (sponge) and 55\% (towel), compared to the baseline's 60\% and 45\%. The performance gap was larger for unseen objects, where our model succeeded 65\% (sponge) and 55\% (towel) of the time, while the baseline dropped to 55\% and 30\%, respectively.
This improvement comes from our model’s ability to choose the most feasible strategy in a given situation. Figure~\ref{fig:rl_demo} showcases planning and action generation on unseen tasks. In towel placement (a), the model prefers a three-step plan—Fold towel → Put in cabinet → Close cabinet—because it predicts a higher success value than placing it directly. In sponge placement (b), with a clock blocking the cabinet, it first pushes the clock aside before placing the sponge, again selecting the most feasible path. System 1 then executes the chosen sequence to produce the shown trajectories, demonstrating practical effectiveness on a physical robot.

%% file: sections/6.conclusion.tex
\section{Discussion and Limitations}
In this section, we discuss the broader implications of our findings, outline current limitations, and present a simple test-time replanning extension that partially mitigates one of these limitations. Finally, we summarize the key contributions and insights drawn from this study.

\subsection{Discussion}
Our experiments demonstrate that incorporating failure information and scaling reasoning at inference significantly enhances the robustness of VLA planning. The observed improvements suggest that failure-aware reasoning quality plays a central role in generalization under distribution shift. By explicitly evaluating and pruning failure-prone branches, the planner learns to allocate computation adaptively, leading to more stable performance across diverse environments. These results also indicate that inference-time computation can serve as a controllable variable for deployment-time adaptation. However, several challenges remain, particularly in integrating real-time feedback and bridging the gap between discrete reasoning and continuous control, which we discuss next.

\subsection{Limitations}
Despite the advantages, we have three main limitations.
First, planning occurs at subgoal granularity without real-time, within-chunk feedback. We deliberately avoid a stop-and-replan loop because it effectively enforces a quasi-static assumption and interrupts continuous motion; in dynamic scenes, small pose errors, sensor latency, or contact disturbances can therefore accumulate between replans, leading to drift in the predicted successor state and occasional divergence between the planned path and what the robot can stably execute.

Second, the low-level controller (System~1) is trained solely from successful executions. Lacking failure-conditioned gradients, it tends to overfit nominal trajectories and shows limited ability to self-correct once perturbed off the known distribution. 
Third, the state abstraction relies on a 2D scene graph that omits metric geometry, contact cues, and short-horizon dynamics that are critical in contact-rich manipulation. This can miscalibrate feasibility estimates when objects are partially occluded, slightly mislocalized, or when frictional effects dominate. The resulting gaps increase sensitivity to calibration drift and viewpoint changes, and can cause the search to misrank otherwise viable branches or prune plans that would succeed under the true physical dynamics.

\subsection{Plug-in Module for Replanning}\label{sec:diss_replan}
\begin{figure}
    \centering
    \includegraphics[width=0.45\linewidth]{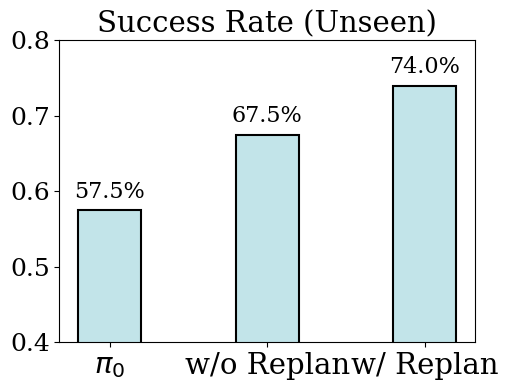}
    \caption{
    Success rates in the unseen drawer packing environment with and without replanning. 
    A lightweight uncertainty-triggered replanning module, which down-weights uncertain branches and switches to alternatives, boosts success in drawer packing without retraining.
    }
    \label{fig:replan}
\end{figure}

\begin{figure}[!t]
    \centering
    \includegraphics[width=0.9\linewidth]{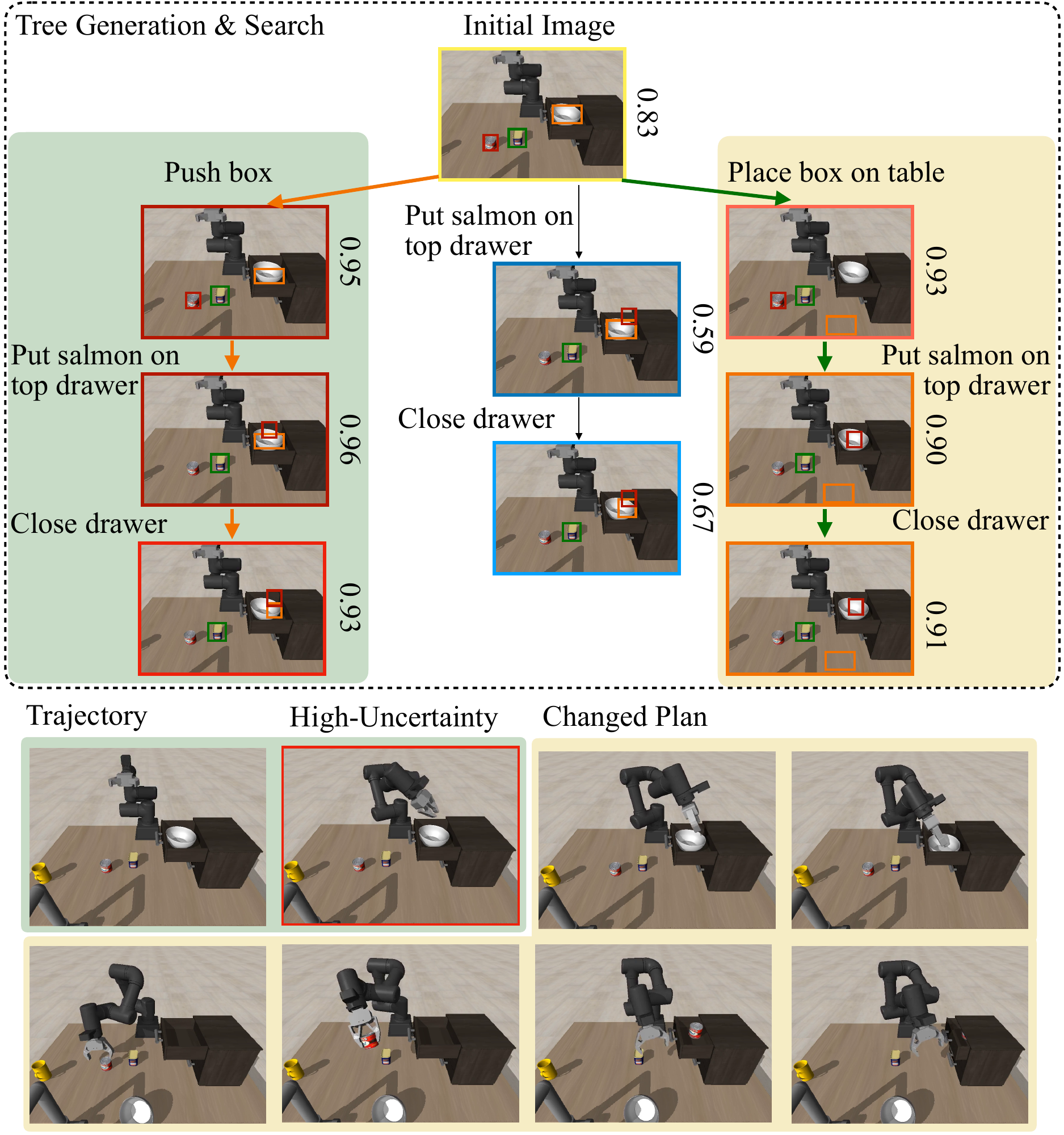}
    \caption{Qualitative Results for replanning. Planner first selects the green plan (\emph{push box}→\emph{place item}→\emph{close}), then detects high uncertainty during the first step and switches to the yellow alternative (\emph{place box on table}→\emph{place item}→\emph{close}), successfully completing the task. Shaded panels indicate the uncertainty trigger and the plan switch; bottom rows show the executed trajectories.}
    \label{fig:replan_demo}
\end{figure}
To mitigate drift arising from executing long action–chunks without feedback, we attach a lightweight uncertainty monitor~\cite{agia2024unpacking} that enables \emph{test-time replanning} without retraining. When the within–chunk variance becomes high, the monitor down-weights the score of currently committed branch and triggers a switch to the next best alternative in the search tree. This converts our quasi-static, open-loop execution into an instance-adaptive procedure that can abandon deteriorating motions. In the unseen drawer packing setting, this mechanism improves success from $67.5\%$ (no replanning) to $74.0\%$ (with replanning), exceeding the $\pi_0$ baseline’s $57.5\%$ while leaving all learned parameters unchanged (Figure~\ref{fig:replan}). Although task-specific calibration remains, these results indicate that uncertainty-triggered replanning partially closes the gap introduced by Limitation~1.
Figure~\ref{fig:replan_demo} illustrates test-time replanning, where the model first chooses the green path—pushing the obstructive box aside—but, upon detecting high uncertainty, it switches (yellow background) to the second-best plan: placing the box on the table. 
Despite the benefits, the current uncertainty calibration is manual and task-specific, and the added computation limits real-time use. We view data-driven calibration and learned failure signals (e.g., extending the done-expert dimension) as promising next steps. These replanning results are orthogonal to our main contributions but suggest a practical path to improve robustness at test time.

\subsection{Conclusion}
In this paper, we introduced \method, a hierarchical vision language action model that provides a framework for effectively training Vision-Language-Action (VLA) models by leveraging both success and failure demonstration data.
By formulating the problem as hierarchical reinforcement learning (HRL), \method separates a high-level system (System 2) for feasibility-aware planning from a low-level control system (System 1) for execution. Crucially, the high-level planner learns from both success and failure examples in offline datasets to guide a tree-based search, allowing it to anticipate and prune failure-prone action sequences before execution, thereby significantly enhancing stability. Through experiments on complex, long-horizon manipulation tasks in both simulation and the real world, \method consistently demonstrates superior success rates and robustness over strong VLA baselines, with its effectiveness being particularly pronounced in unseen scenarios. These results underscore that explicitly modeling and reasoning about failure is an essential component for converting the broad competence of VLA models into robust real-world performance.

%% file: sections/appendix.tex
\appendix

\subsection{Detailed Formulation}
\subsubsection{Node-edge kernel induced by low-level options}
We define the high-level node-edge space from the low-level MDP $(\mathcal S,\mathcal A,T)$ and show that it forms an MDP, and we introduce the node-level target/avoid sets corresponding to $\mathcal G,\mathcal F\subset\mathcal S$.

Given the option grounding $o_e=(I_e^{N},\,\pi_e,\,\beta_e,\,\Pi_e)$, the closed-loop one-step kernel under $o_e$ is
\[
P_{\pi_e}(s'\mid s)=\sum_{a\in\mathcal A}\pi_e(a\mid s,\ell)\,T(s'\mid s,a).
\]
With termination-on-arrival, the terminal low-level state distribution is
\[
P^{S}_{o_e}(s^\star\mid s)=\sum_{t\ge 0}\big((1-\beta_e)P_{\pi_e}\big)^{t}\,(\beta_e P_{\pi_e})(s^\star\mid s),
\]
i.e., the probability that, starting from $s$ and executing $o_e$, the execution eventually terminates in $s^\star$ under $\beta_e$.

Lift the high-level pair $(n,z)$ to a low-level \emph{arrival} distribution $\mu(\cdot\mid n,z)$, the conditional law of $s$ upon arriving at node $n$ after the option sequence encoded in $z$ (in data, $\mu=\delta_{s_{t_k}}$ with $\phi(s_{t_k})=n$). Mapping terminal states to nodes via $\Pi_e:\mathcal S\!\to\!\mathcal N$ yields the \emph{node-level option kernel with context}
\[
P^{N}_{o_e}(n'\mid n,z)
=\mathbb E_{s\sim \mu(\cdot\mid n,z)}
\Big[\sum_{s^\star} P^{S}_{o_e}(s^\star\mid s)\,\mathbf 1\{\Pi_e(s^\star)=n'\}\Big].
\]
With the deterministic context update $z'=[z,\,e,\,n]$, the \emph{node-edge} kernel is
\[
P\big((n',z')\mid(n,z),e\big)=P^{N}_{o_e}(n'\mid n,z)\;\mathbf 1\{\,z'=[z,e,n]\,\}.
\]
Define the node-level target/avoid sets by lifting through $\phi$:
\[
\mathcal G_N:=\phi(\mathcal G),\qquad \mathcal F_N:=\phi(\mathcal F),\qquad \mathcal G_N\cap\mathcal F_N=\varnothing.
\]
The kernel is absorbing on $\mathcal G_N\cup\mathcal F_N$. Therefore $(\mathcal X,\mathcal U,P)$ with $\mathcal X=\mathcal N\times\mathcal Z$, $\mathcal U=\mathcal E$, and transition kernel $P$ is a well-defined MDP on the node-edge space.

\subsubsection{Value equals reachability probability (undiscounted first exit)}
Define decision-epoch hitting times
\begin{multline}
    \tau_{\mathcal G}:=\inf\{t\ge 0:\ n_{t}\in\mathcal G_N\},\qquad \\
    \tau_{\mathcal F}:=\inf\{t\ge 0:\ n_{t}\in\mathcal F_N\},\qquad \\
    \tau:=\tau_{\mathcal G}\wedge\tau_{\mathcal F}.
\end{multline}
With the entrance reward
$r(n,z,e,n')=\mathbf 1\{n'\in\mathcal G_N\}$
and \emph{undiscounted} first–exit return
\[
R \;:=\; \sum_{t=0}^{\tau-1} r(n_t,z_t,e_t,n_{t+1}),
\]
we have the pathwise identity
\[
R \;=\; \mathbf 1\{\tau_{\mathcal G}<\tau_{\mathcal F}\}.
\]
Indeed, exactly one nonzero reward can occur—on the unique transition that first enters
$\mathcal G_N$ (if any); if $\mathcal F_N$ is hit first, all rewards are zero and the
episode terminates.

For any policy $\mu$ on the node–edge MDP $(\mathcal X,\mathcal U,P)$ with
$\mathcal X=\mathcal N\times\mathcal Z$, $\mathcal U=\mathcal E$, and kernel
$P((n',z')\mid(n,z),e)$,
\begin{multline}
V^\mu(n,z)
\;:=\;
\mathbb E_\mu\!\left[\sum_{t=0}^{\tau-1} r(n_t,z_t,e_t,n_{t+1})
\;\middle|\; (n_0,z_0)=(n,z)\right] \\
\;=\;
\Pr_\mu\!\big(\tau_{\mathcal G}<\tau_{\mathcal F}\mid n,z\big).
\end{multline}

The corresponding Bellman relations are the standard reach-avoid equations:
\begin{multline}
V^\mu(n,z)=\sum_{e}\mu(e\mid n,z)\!\!\sum_{n'}\!P^{N}_{o_e}(n'\mid n,z)
\Bigl[\mathbf 1\{n'\in\mathcal G_N\}   \\
+\mathbf 1\{n'\notin \mathcal G_N\cup\mathcal F_N\}V^\mu(n', [z,e,n])\Bigr],    
\end{multline}

with boundary conditions $V^\mu(n,z)=1$ for $n\in\mathcal G_N$ and $V^\mu(n,z)=0$ for $n\in\mathcal F_N$.
The optimality version replaces the inner expectation by a maximization over $e\in\mathcal O(n,z)$.
If a discount factor \(\gamma\in(0,1)\) is preferred, define \(R_\gamma=\sum_{t=0}^{\tau-1}\gamma^{t} r(n_t,z_t,e_t,n_{t+1})\) and note that pathwise \(R_\gamma=\gamma^{\tau_{\mathcal G}-1}\mathbf 1\{\tau_{\mathcal G}<\tau_{\mathcal F}\}\) so
\begin{multline}
V^\mu_\gamma(n,z)
=\mathbb E_\mu[R_\gamma\mid n,z] \\
=\Pr_\mu(\tau_{\mathcal G}<\tau_{\mathcal F}\mid n,z)\, 
\mathbb{E}_\mu\!\big[\gamma^{\tau_{\mathcal G}-1}\mid \tau_{\mathcal G}<\tau_{\mathcal F},n,z\big].
\end{multline} 
The discounted value is therefore proportional to the success probability with a positive factor depending on the success time; when search evaluates leaves at an approximately fixed depth this factor is effectively constant so ranking by \(V^\mu_\gamma\) coincides with ranking by reachability, and if a probability scale is required one may normalize by depth or divide by an empirical estimate of the factor computed from the dataset.

\subsubsection{Expectile losses on the node--edge MDP}
Let $\mathcal D=\{(n_k,z_{0:k-1},e_k,n_{k+1})\}$ be logged transitions from a behavior policy $\mu_b(e\mid n,z)$.
Define the entrance reward $r_k=\mathbf 1\{n_{k+1}\in\mathcal G_N\}$ and the terminal flag
$\mathrm{term}_k=\mathbf 1\{n_{k+1}\in\mathcal G_N\cup\mathcal F_N\}$.
Write $\theta$ for trainable parameters and $\theta'$ for an EMA target.

From the Bellman equation under $\mu_b$,
\[
V^{\mu_b}(n_k,z_{0:k-1})=\mathbb E\!\left[r_k+\gamma(1-\mathrm{term}_k)\,V^{\mu_b}(n_{k+1},z_{0:k})\right],
\]
we form the one-step target
\[
y_k \;=\; r_k \;+\; \gamma(1-\mathrm{term}_k)\,V_{\theta'}(n_{k+1},z_{0:k}).
\]
We then minimize the expectile–TD loss
\begin{multline}
    \mathcal L_{\mathrm{val}}(\theta)
=\mathbb E_{\mathcal D}\!\left[
\rho_{\tau_e}\!\big(y_k - V_\theta(n_k,z_{0:k-1})\big)
\right],\\
\rho_{\tau}(u)=\big|\tau-\mathbf 1\{u<0\}\big|\,u^2.
\end{multline}

While original TD with symmetric MSE is tailored to online RL where fresh on-policy samples expand coverage and the Bellman contraction drives $V_{\theta}$ toward the true value as data grows, our setting is offline. 
We train on a fixed batch, so TD error is reduced only on the logged support, and bootstrap propagation outside that support can induce overestimation on unseen actions or edges. 
To guard against this, we adopt the expectile TD loss with a pessimistic level $\tau_e<0.5$, which penalizes positive residuals $V_\theta-y_k$ more than negative ones and yields a conservative value on under-supported regions. The objective remains a simple and stable TD regression that avoids importance weighting. With $\gamma=1$ and first-exit rewards, $V_\theta(n,z)$ approximates $\Pr(\tau_{\mathcal G}<\tau_{\mathcal F}\mid n,z)$, and $\tau_e$ provides a clean knob to trade off recall and caution during planning under offline data constraints.

\subsection{Dataset Details}
\subsubsection{Plug Insertion Environment}
\begin{figure}
    \centering
    \includegraphics[width=0.9\linewidth]{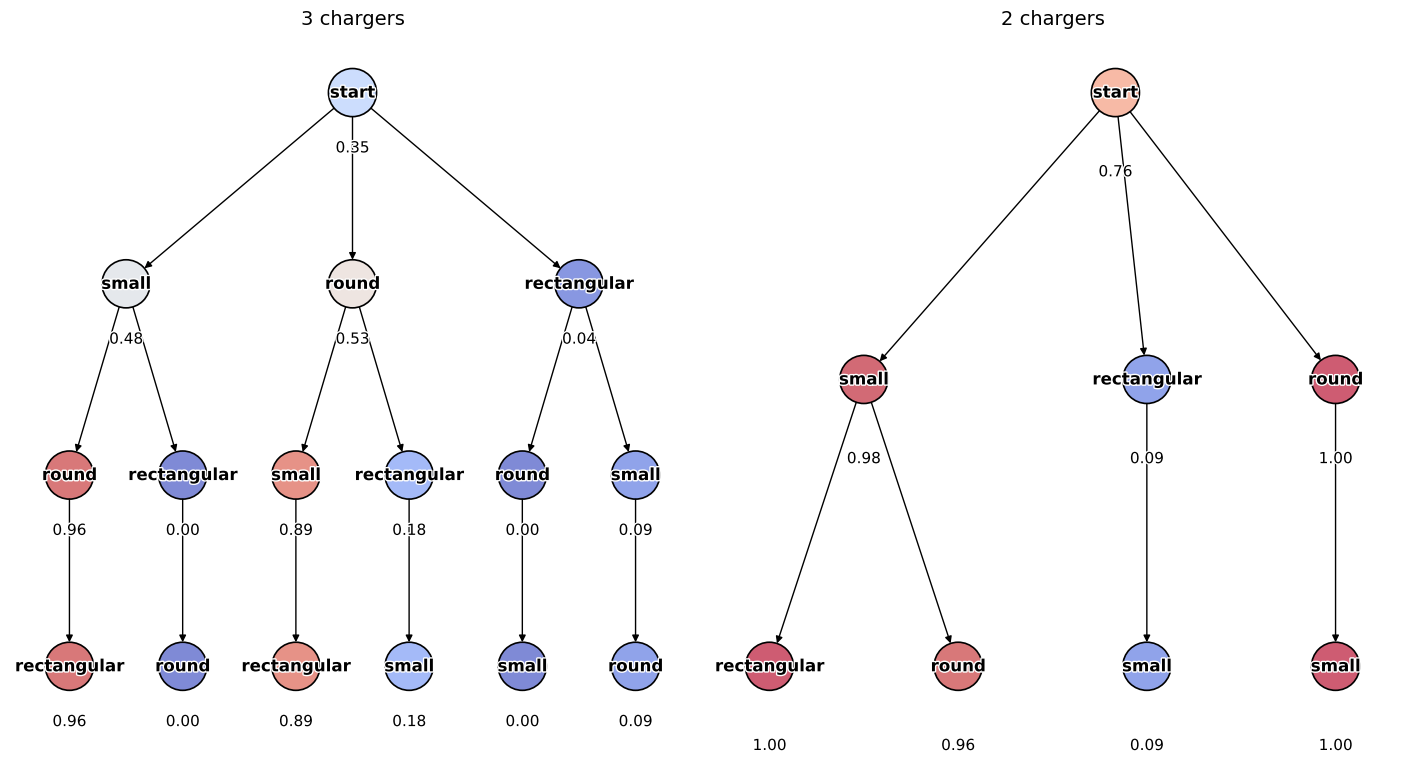}
    \caption{Data Statistics from human demonstrations in plug insertion environment}
    \label{fig:data}
\end{figure}

\paragraph{Collection.}
We used teleoperation with four operators to cover \emph{all admissible insertion orders}
for each receptacle type (e.g., all permutations for 3--socket plates and for 2--socket plates). The insertion order for strips is fixed from right to left due to limited viewpoints.
Operators received order guidelines beforehand so each trajectory explicitly followed
a target sequence (e.g., \texttt{small} $\rightarrow$ \texttt{round} $\rightarrow$ \texttt{rectangular}).
For every run we logged time--stamped robot state (arm pose, gripper command), RGB images,
and camera calibration.

\paragraph{Subgoal labels.}
From the logs we derive a minimal set of discrete subgoals tied to gripper state transitions
and the target order:
\begin{itemize}
  \item \textbf{Grasp-$x$:} \texttt{open} $\rightarrow$ \texttt{close} near plug $x$,
  \item \textbf{Insert-$x$ in $y$:} \texttt{close} $\rightarrow$ \texttt{open} at socket $y$.
\end{itemize}
A trajectory thus becomes a chain of subgoals aligned to the instructed order
(e.g., \texttt{grasp-small charger} $\rightarrow$ \texttt{insert-small charger in rightmost plug} $\rightarrow$ \texttt{grasp-round charger} $\rightarrow \cdots$).

\paragraph{Scene snapshots.}
Around each transition time, we extract a short image window and select the central
frame as the \emph{scene} for that subgoal.

\paragraph{Scene graph construction.}
For every scene, we build a 2D scene graph containing:
\begin{itemize}
  \item \textbf{Gripper node:} 2D image location obtained by projecting the measured 3D gripper pose
        into the camera frame using known intrinsics/extrinsics.
  \item \textbf{Object nodes:} bounding--box proposals from \emph{GroundingDINO~\cite{liu2024grounding}}; a VLM (\emph{Gemini~\cite{comanici2025gemini}})
        assigns semantic names (e.g., \texttt{small}, \texttt{round}, \texttt{rectangular}) to crops via text prompts seeded by the task vocabulary.
  \item \textbf{Relations (edges):} coarse spatial relations between nodes (e.g., charger inside leftmost plug),
        inferred from box geometry and, when ambiguous, VLM judgments.
\end{itemize}

\paragraph{Output triplets.}
Each subgoal yields a \{image, scene-graph, label\} triple: the raw image,
a graph with the gripper 2D position, object boxes, relations, and the symbolic subgoal induced by the gripper transition and the instructed order.

\paragraph{Data Statistics}
Figure~\ref{fig:data} visualizes the empirical \emph{per–path success rate}
for the plug–insertion task from human demonstrations. Each tree enumerates all candidate insertion paths
(from the root \texttt{start} to a leaf), where a leaf corresponds to a \emph{complete}
order of socket types (\texttt{small}, \texttt{round}, \texttt{rectangular}).
The number under each leaf is the estimated success probability
$\widehat{p}_{\text{succ}}=\sharp \text{successes} / \sharp \text{trials}$ for that path, and
node color encodes this value (red $\rightarrow$ high, blue $\rightarrow$ low).
The left panel reports the 3-charger setting and the right panel the 2-charger setting.
The distribution is skewed: in 3–3-charger, sequences such as
\texttt{small} $\rightarrow$ \texttt{round} $\rightarrow$ \texttt{rectangular}
are highly reliable ($0.96$), while any path involving \texttt{rectangular} early
tends to fail ($0.00$–$0.18$). In a 2-charger, several paths are near–perfect
(e.g., \texttt{round} $\rightarrow$ \texttt{small} and
\texttt{small} $\rightarrow$ \texttt{rectangular} both at $1.00$),
whereas \texttt{rectangular} $\rightarrow$ \texttt{small} is weak ($0.09$).

\subsubsection{Drawer Packing Environment}

\begin{figure}
    \centering
    \includegraphics[width=0.9\linewidth]{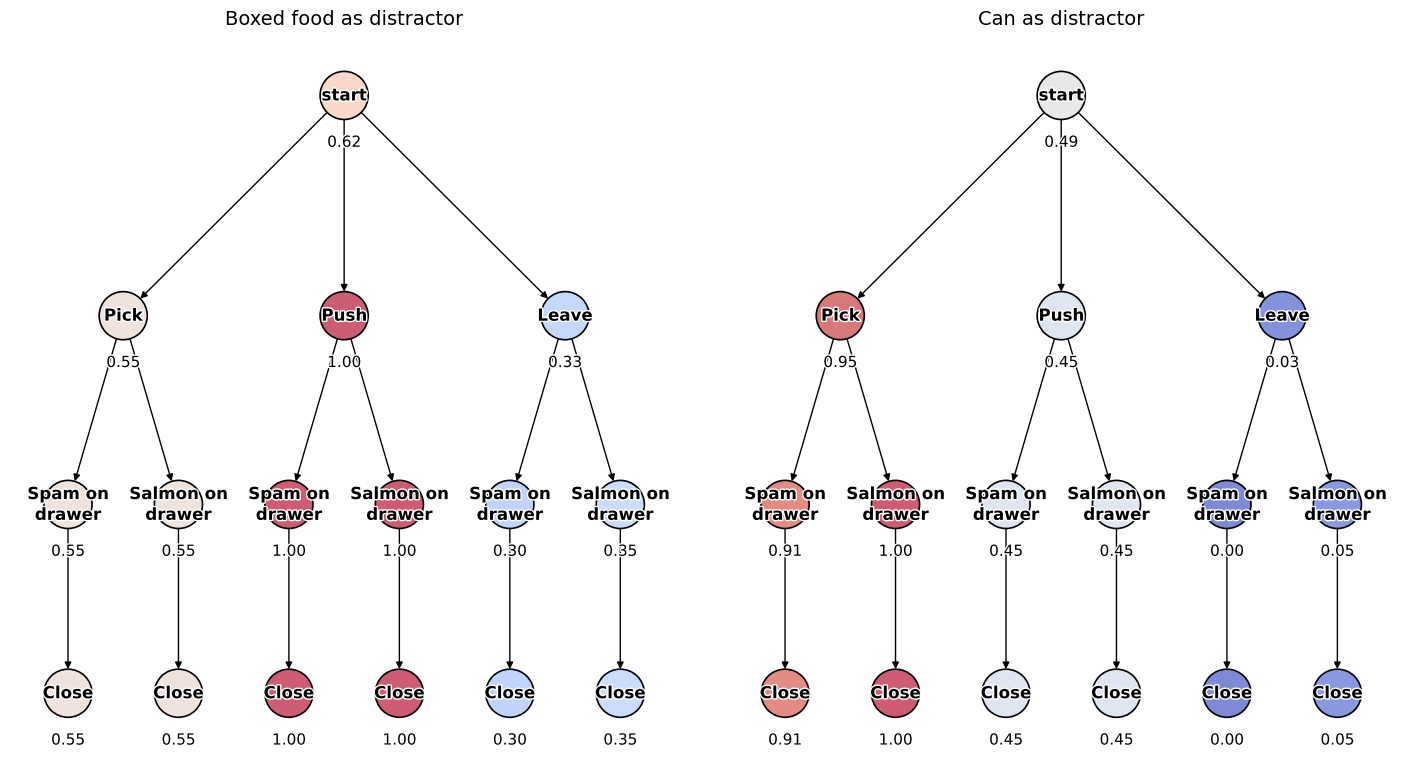}
    \caption{Data Statistics from human demonstrations in the drawer packing environment}
    \label{fig:data_drawer}
\end{figure}
\paragraph{Collection.}
We used teleoperation with a leader–follower arm setup to collect demonstrations of
drawer packing. Four operators performed trajectories covering all admissible
object–placement orders across different categories (\texttt{boxed food}, \texttt{cans}, etc.).
For each run, the teleoperator followed a target guideline (e.g., \texttt{pick box $\rightarrow$ put on table $\rightarrow$ grasp canned spam $\rightarrow$ put in drawer}) ensuring coverage of \emph{Pick}, \emph{Push}, and \emph{Leave} strategies.
The initial scene includes two types of distractors—\texttt{boxed food} items and \texttt{cans} placed on (or near) the drawer surface.
The task, conditioned on the language instruction, is to select the specified can (\texttt{spam} or \texttt{salmon}), place it inside the drawer, and then close the drawer.
The dataset logs include time–stamped robot states (joint angles, gripper signal),
RGB observations, object poses from the MuJoCo simulator, and synchronized language instructions.

\paragraph{Subgoal labels.}
From the logs we derive a minimal set of discrete subgoals aligned with gripper state
transitions and the target order. Each trajectory is segmented into a chain of symbolic steps:
\begin{itemize}
  \item \textbf{Pick-$x$:} grasp and lift a distractor object $x$ from the table,
  \item \textbf{Push-$x$:} displace object $x$ away to clear space,
  \item \textbf{Grasp-$y$:} grasp the target food item $y$ (e.g., canned spam, salmon),
  \item \textbf{Put-$y$ on drawer:} place $y$ into the top drawer,
  \item \textbf{Close drawer:} close the top drawer with the gripper.
\end{itemize}
Thus, each demonstration becomes an ordered sequence of subgoals (e.g.,
\texttt{push box $\rightarrow$ grasp spam $\rightarrow$ put spam on drawer $\rightarrow$ close drawer}),
which provides the symbolic scaffold for scene–graph based reasoning.

Scene graphs are constructed in the same method as a plug insertion environment.

\paragraph{Data Statistics}
Figure~\ref{fig:data_drawer} illustrates the empirical \emph{per–path success rate} in the drawer–packing task. Each tree expands all candidate sequences from the root \texttt{start} to a leaf, where a leaf specifies a complete decision (e.g., placing \texttt{Spam} or \texttt{Salmon}). The numeric value under each node is the estimated success probability $\widehat{p}_{\text{succ}}=\sharp\text{successes} / \sharp\text{trials}$, and the node color encodes this value (red $\rightarrow$ high, blue $\rightarrow$ low).

In the \texttt{boxed food} setting (left), the distribution is uneven: \texttt{Push} actions are consistently reliable ($1.00$ for both items), while \texttt{Pick} yields moderate success ($0.55$) and \texttt{Leave} is weak ($0.30$–$0.35$). In contrast, the \texttt{can} setting (right) shows a sharper split: \texttt{Pick} is highly reliable ($0.91$–$1.00$), \texttt{Push} succeeds less often ($0.45$), and \texttt{Leave} almost always fails ($0.00$–$0.05$). This highlights how different object types bias subgoal feasibility—pushing works well for boxed food, while picking dominates for cans.

\subsubsection{Simpler Environment}
\paragraph{Collection and Setup.}
In this environment, we replace teleoperation with autonomous rollouts from a fine-tuned \(\pi_0\) policy. Each trajectory is labeled with a binary success signal from the environment. We segment behavior into two edge types parameterized by the 2D gripper position: \textbf{grasp\(-x\)} and \textbf{put \(x\) on top of \(y\)}. The grasping interval is annotated using the environment log flag \texttt{info["is\_src\_obj\_grasped"]}.

\paragraph{Data Statistics.}
We collected \(576\) trajectories across four objects: \texttt{eggplant} (72), \texttt{carrot} (192), \texttt{spoon} (120), and \texttt{cube} (192). Table~\ref{tab:simpler_stats} summarizes per-object grasp and success outcomes.

\begin{table}[h]
\centering
\caption{Simpler environment outcomes per object. Rates are $\sharp$ / total (\%).}
\label{tab:simpler_stats}
\setlength{\tabcolsep}{6pt}
\begin{tabular}{lrrrr}
\toprule
Object & Total & Grasped & Success \\
\midrule
eggplant & 72  & 66 \;(91.7\%) & 60 \;(83.3\%) \\
carrot   & 192 & 112 (58.3\%) & 48 \;(25.0\%) \\
spoon    & 144 & 77 \;(53.4\%) & 56 \;(38.8\%) \\
cube     & 192 & 128 (66.7\%) & 32 \;(16.7\%) \\
\midrule
\textbf{Overall} & \textbf{600} & \textbf{376 (65.3\%)} & \textbf{195 (33.9\%)} \\
\bottomrule
\end{tabular}
\end{table}

\subsubsection{Real-World Environment}
\begin{figure}
    \centering
    \includegraphics[width=0.9\linewidth]{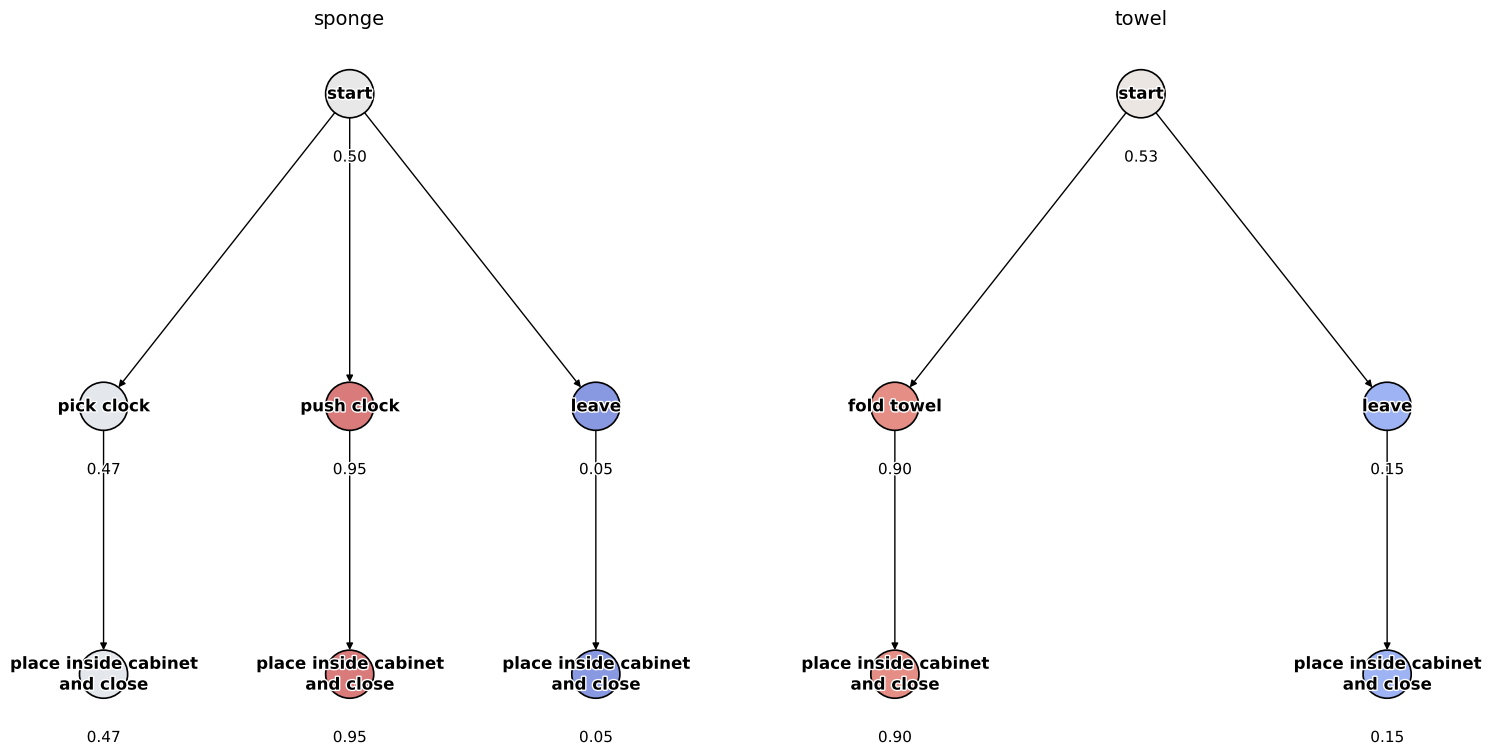}
    \caption{Data Statistics from human demonstrations in the real-world environment}
    \label{fig:data_real}
\end{figure}

\paragraph{Collection and Setup.}
We study a cabinet–packing task where the agent must place either a \texttt{sponge} or a \texttt{towel} into a lidded cabinet bin and close it. For \texttt{sponge} trials, a \texttt{clock} distractor is present inside the bin; operators could (i) \emph{push} the clock aside, (ii) \emph{pick} it up and \emph{place} it elsewhere, or (iii) \emph{leave} it and insert the sponge. For \texttt{towel} trials, operators either \emph{fold once} before insertion or \emph{insert as is}. A rollout is marked as successful if the lid fully closes with no protrusion. We used an OpenManipulator-Y arm under leader–follower teleoperation, logging at $20$HZ. 

\paragraph{Subgoal labels.}
From the time–stamped logs (20 Hz) we derive a minimal set of discrete subgoals aligned with gripper state transitions and the target order. Each trajectory is segmented into a chain of symbolic steps:
\begin{itemize}
\item \textbf{Pick-$x$:} grasp and lift a distractor $x$ (e.g., the \texttt{clock}) and relocate it,
\item \textbf{Push-$x$:} displace $x$ to clear space without grasping,
\item \textbf{Fold-\textit{towel}:} fold the towel once prior to insertion (towel trials only),
\item \textbf{Place $y$ in cabinet:} insert the target item $y$ (e.g., \texttt{sponge}, \texttt{towel}) into the cabinet bin,
\item \textbf{Close cabinet:} close the lid with no protrusions.
\end{itemize}
The \emph{leave} strategy denotes omitting any distractor–handling step. Thus, each demonstration becomes an ordered subgoal sequence, e.g.,
\texttt{push clock $\rightarrow$ place sponge in cabinet $\rightarrow$ close cabinet} (sponge–push),
\texttt{pick clock $\rightarrow$ place sponge in cabinet $\rightarrow$ close cabinet} (sponge–pick),
\texttt{place sponge in cabinet $\rightarrow$ close cabinet} (sponge–leave),
\texttt{fold towel $\rightarrow$ place towel in cabinet $\rightarrow$ close cabinet} (towel–fold), or
\texttt{place towel in cabinet $\rightarrow$ close cabinet} (towel–leave),
which provides the symbolic scaffold for downstream scene–graph–based reasoning and analysis.
\paragraph{Data Statistics.}
Figure~\ref {fig:data_real} visualizes the empirical \emph{per–path success rate} for the cabinet–insertion task, with each root-to-leaf path representing a complete subgoal sequence. The numeric label under each node is the estimated success probability $\widehat{p}_{\text{succ}}=\sharp \text{successes} / \sharp \text{trials}$, and node color encodes this value.

In \texttt{sponge} trials (with an internal \texttt{clock} distractor), \textbf{Push-clock} dominates ($0.95$, $20/21$), \textbf{Pick-clock} is moderate ($0.47$, $9/19$), and \textbf{Leave} almost always fails ($0.05$, $1/20$). In \texttt{towel} trials, \textbf{Fold} before insertion is highly reliable ($0.90$, $18/20$), whereas \textbf{Leave} is weak ($0.15$, $3/20$). Overall, handling the distractor (push/pick) or preparing the deformable item (fold) is crucial for success, while omitting these steps yields poor outcomes.



\subsection{Ablation Studies}

\begin{figure}[!t]
    \centering
    \includegraphics[width=0.9\linewidth]{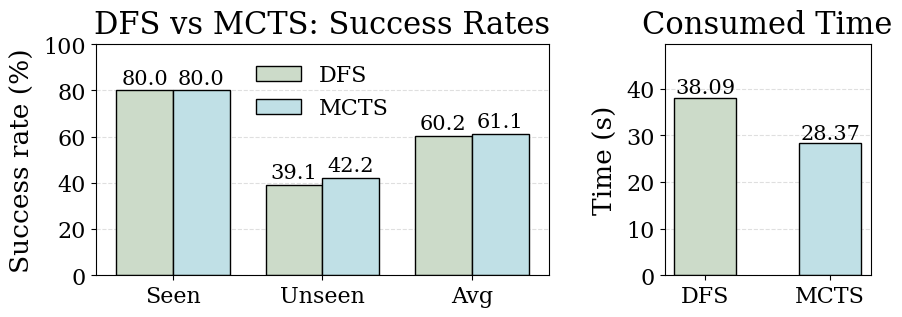}
    \caption{Comparison between tree search algorithms. On plug insertion, MCTS’s value-guided search yields a similar success rate and faster inference than fixed-order DFS.}
    \label{fig:dfs}
\end{figure}


\paragraph{Tree Search Algorithm}
To validate our choice of a Monte Carlo Tree Search (MCTS)-like algorithm, we compare it against a standard Depth-First Search (DFS) baseline in the plug insertion task (Figure \ref{fig:dfs}). While both methods achieve an identical 80.0\% success rate in the seen setting, their performance diverges in the more challenging unseen setting. Here, our MCTS-like approach achieves a higher success rate (42.2\% vs. 39.1\% for DFS). More importantly, it is significantly more computationally efficient, reducing the average execution time by nearly 10 seconds (28.37s vs. 38.09s for DFS). This efficiency stems from MCTS's guided exploration, which uses the value function to prioritize more promising branches. In contrast, DFS explores paths in a fixed order, often wasting computation on unfeasible plans. This confirms our search strategy is not only more effective at generalizing but also makes better use of computational resources.

\begin{table}[!t]
    \centering
    \caption{Weight merging for language backbone in plug insertion environment. Interpolating shared layers between \(\pi_{0}\) and PaliGemma via \(\lambda_m\) yields a clear optimum at \(\lambda_m=0.6\); heavier bias to either model reduces overall and unseen success.}
    \begin{tabular}{c|ccc}
      \toprule
      $\lambda_m$ & Seen SR. & Unseen SR. & Avg. SR. \\
      \midrule
      0.5 & 0.756 & 0.356 & 0.555 \\
      \rowcolor{blue!10} 0.6 & \textbf{0.800} & \textbf{0.422} & \textbf{0.611} \\
      0.7 & 0.733 & 0.289 & 0.511 \\
      0.8 & 0.422 & 0.244 & 0.333 \\
      \bottomrule
    \end{tabular}%
    \label{tab:weight_merge}
\end{table}

\paragraph{Weight Merging of Language Backbone}
We ablate backbone weight merging~\cite{wortsman2022model} between PaliGemma~\cite{beyer2024paligemma} and the $\pi_{0}$~\cite{black2024pi_0} trunk. Specifically, we linearly interpolate the shared layers,
\[
\theta^{\mathrm{merge}}_{g}(\lambda_m)=\lambda_m\,\theta^{\pi_{0}}_{g}+(1-\lambda_m)\,\theta^{\mathrm{PG}}_{g},
\]
while keeping non-overlapping heads separate and training LoRA~\cite{hu2022lora} on top.

Table~\ref{tab:weight_merge} shows the effect of varying interpolation weight $\lambda$. 
Performance peaks at $\lambda_m{=}0.6$ , achieving our best results with a seen success rate of 0.800, an unseen success rate of 0.422, and an average of 0.611. Decreasing the weight to $\lambda_m{=}0.5$  (giving more weight to PaliGemma) maintains a high seen success rate (0.756) but harms generalization, with the unseen rate dropping to 0.356. Conversely, increasing the weight towards $\pi_0$ with $\lambda_m{=}0.7$  (Seen: 0.733, Unseen: 0.289) and $\lambda_m{=}0.8$  (Seen: 0.422, Unseen: 0.244) leads to a steep decline in performance across all metrics. This highlights the importance of finding a careful balance between the general linguistic knowledge from PaliGemma and the specialized action representations from the $\pi_0$ trunk to achieve both strong in-distribution performance and robust generalization.

%% file: reference.bib
@inproceedings{
yuan2023rewarddirected,
title={Reward-Directed Conditional Diffusion: Provable Distribution Estimation and Reward Improvement},
author={Hui Yuan and Kaixuan Huang and Chengzhuo Ni and Minshuo Chen and Mengdi Wang},
booktitle={Proc. of the Thirty-seventh Conference on Neural Information Processing Systems (NeurIPS)},
year={2023},
url={https://openreview.net/forum?id=58HwnnEdtF}
}

@inproceedings{brohan2022rt,
  title={Rt-1: Robotics transformer for real-world control at scale},
  author={Brohan, Anthony and Brown, Noah and Carbajal, Justice and Chebotar, Yevgen and Dabis, Joseph and Finn, Chelsea and Gopalakrishnan, Keerthana and Hausman, Karol and Herzog, Alex and Hsu, Jasmine and others},
  booktitle={Proc. of the Robotics: Science and System (RSS)},
  year={2023}
}

@inproceedings{octo_2023,
    title={Octo: An Open-Source Generalist Robot Policy},
    author = {{Octo Model Team} and Dibya Ghosh and Homer Walke and Karl Pertsch and Kevin Black and Oier Mees and Sudeep Dasari and Joey Hejna and Charles Xu and Jianlan Luo and Tobias Kreiman and {You Liang} Tan and Lawrence Yunliang Chen and Pannag Sanketi and Quan Vuong and Ted Xiao and Dorsa Sadigh and Chelsea Finn and Sergey Levine},
    booktitle = {Proc. of the Robotics: Science and Systems (RSS)},
    address  = {Delft, Netherlands},
    year = {2024},
}

@inproceedings{rt2_2023,
  title = 	 {RT-2: Vision-Language-Action Models Transfer Web Knowledge to Robotic Control},
  author =       {Zitkovich, Brianna and Yu, Tianhe and Xu, Sichun and Xu, Peng and Xiao, Ted and Xia, Fei and Wu, Jialin and Wohlhart, Paul and Welker, Stefan and Wahid, Ayzaan and Vuong, Quan and Vanhoucke, Vincent and Tran, Huong and Soricut, Radu and Singh, Anikait and Singh, Jaspiar and Sermanet, Pierre and Sanketi, Pannag R. and Salazar, Grecia and Ryoo, Michael S. and Reymann, Krista and Rao, Kanishka and Pertsch, Karl and Mordatch, Igor and Michalewski, Henryk and Lu, Yao and Levine, Sergey and Lee, Lisa and Lee, Tsang-Wei Edward and Leal, Isabel and Kuang, Yuheng and Kalashnikov, Dmitry and Julian, Ryan and Joshi, Nikhil J. and Irpan, Alex and Ichter, Brian and Hsu, Jasmine and Herzog, Alexander and Hausman, Karol and Gopalakrishnan, Keerthana and Fu, Chuyuan and Florence, Pete and Finn, Chelsea and Dubey, Kumar Avinava and Driess, Danny and Ding, Tianli and Choromanski, Krzysztof Marcin and Chen, Xi and Chebotar, Yevgen and Carbajal, Justice and Brown, Noah and Brohan, Anthony and Arenas, Montserrat Gonzalez and Han, Kehang},
  booktitle = 	 {Proc. of The 7th Conference on Robot Learning (CoRL)},
  pages = 	 {2165-2183},
  year = 	 {2023},
}

@misc{rt_x_2023,
title={Open {X-E}mbodiment: Robotic Learning Datasets and {RT-X} Models},
author = {Open X-Embodiment Collaboration },
howpublished  = {\url{https://arxiv.org/abs/2310.08864}},
year = {2023},
}

@inproceedings{
kim2024openvla,
title={Open{VLA}: An Open-Source Vision-Language-Action Model},
author={Moo Jin Kim and Karl Pertsch and Siddharth Karamcheti and Ted Xiao and Ashwin Balakrishna and Suraj Nair and Rafael Rafailov and Ethan P Foster and Pannag R Sanketi and Quan Vuong and Thomas Kollar and Benjamin Burchfiel and Russ Tedrake and Dorsa Sadigh and Sergey Levine and Percy Liang and Chelsea Finn},
booktitle={Proc. of the 8th Annual Conference on Robot Learning (CoRL)},
year={2024},
url={https://openreview.net/forum?id=ZMnD6QZAE6}
}

@article{kim2025fine,
  title={Fine-Tuning Vision-Language-Action Models: Optimizing Speed and Success},
  author={Kim, Moo Jin and Finn, Chelsea and Liang, Percy},
  journal={arXiv preprint arXiv:2502.19645},
  year={2025}
}

@inproceedings{wen2025diffusionvla,
  title={DiffusionVLA: Scaling Robot Foundation Models via Unified Diffusion and Autoregression},
  author={Wen, Junjie and Zhu, Yichen and Zhu, Minjie and Tang, Zhibin and Li, Jinming and Zhou, Zhongyi and Liu, Xiaoyu and Shen, Chaomin and Peng, Yaxin and Feng, Feifei},
  booktitle={Proc. of the Forty-second International Conference on Machine Learning (ICML)},
  year={2025}
}

@article{black2024pi_0,
  title={$\pi_0 $: A Vision-Language-Action Flow Model for General Robot Control},
  author={Black, Kevin and Brown, Noah and Driess, Danny and Esmail, Adnan and Equi, Michael and Finn, Chelsea and Fusai, Niccolo and Groom, Lachy and Hausman, Karol and Ichter, Brian and others},
  journal={arXiv preprint arXiv:2410.24164},
  year={2024}
}

@misc{pi_05_2025,
      title={$\pi_{0.5}$: a Vision-Language-Action Model with Open-World Generalization}, 
      author={Physical Intelligence},
      year={2025},
      eprint={2504.16054},
      archivePrefix={arXiv},
      primaryClass={cs.LG},
      url={https://arxiv.org/abs/2504.16054}, 
}

@article{bjorck2025gr00t,
  title={Gr00t n1: An open foundation model for generalist humanoid robots},
  author={Bjorck, Johan and Casta{\~n}eda, Fernando and Cherniadev, Nikita and Da, Xingye and Ding, Runyu and Fan, Linxi and Fang, Yu and Fox, Dieter and Hu, Fengyuan and Huang, Spencer and others},
  journal={arXiv preprint arXiv:2503.14734},
  year={2025}
}

@article{lbmtri2025,
  title={A Careful Examination of Large Behavior Models for Multitask Dexterous Manipulation}, 
  author={TRI LBM Team},
  year={2025},
  eprint={2507.05331},
  archivePrefix={arXiv},
  primaryClass={cs.RO},
  url={https://arxiv.org/abs/2507.05331}, 
}

@inproceedings{
zawalski2024robotic,
title={Robotic Control via Embodied Chain-of-Thought Reasoning},
author={Micha{\l} Zawalski and William Chen and Karl Pertsch and Oier Mees and Chelsea Finn and Sergey Levine},
booktitle={Proc. of the 8th Annual Conference on Robot Learning (CoRL)},
year={2024},
url={https://openreview.net/forum?id=S70MgnIA0v}
}

@inproceedings{zhao2025cot,
  title={Cot-vla: Visual chain-of-thought reasoning for vision-language-action models},
  author={Zhao, Qingqing and Lu, Yao and Kim, Moo Jin and Fu, Zipeng and Zhang, Zhuoyang and Wu, Yecheng and Li, Zhaoshuo and Ma, Qianli and Han, Song and Finn, Chelsea and others},
  booktitle={Proc. of the Computer Vision and Pattern Recognition Conference (CVPR)},
  pages={1702--1713},
  year={2025}
}

@article{li2024improving,
  title={Improving vision-language-action models via chain-of-affordance},
  author={Li, Jinming and Zhu, Yichen and Tang, Zhibin and Wen, Junjie and Zhu, Minjie and Liu, Xiaoyu and Li, Chengmeng and Cheng, Ran and Peng, Yaxin and Feng, Feifei},
  journal={arXiv preprint arXiv:2412.20451},
  year={2024}
}

@article{huang2025thinkact,
  title={ThinkAct: Vision-Language-Action Reasoning via Reinforced Visual Latent Planning},
  author={Huang, Chi-Pin and Wu, Yueh-Hua and Chen, Min-Hung and Wang, Yu-Chiang Frank and Yang, Fu-En},
  journal={arXiv preprint arXiv:2507.16815},
  year={2025}
}

@article{lee2025molmoact,
  title={MolmoAct: Action Reasoning Models that can Reason in Space},
  author={Lee, Jason and Duan, Jiafei and Fang, Haoquan and Deng, Yuquan and Liu, Shuo and Li, Boyang and Fang, Bohan and Zhang, Jieyu and Wang, Yi Ru and Lee, Sangho and others},
  journal={arXiv preprint arXiv:2508.07917},
  year={2025}
}

@article{shi2025hi,
  title={Hi robot: Open-ended instruction following with hierarchical vision-language-action models},
  author={Shi, Lucy Xiaoyang and Ichter, Brian and Equi, Michael and Ke, Liyiming and Pertsch, Karl and Vuong, Quan and Tanner, James and Walling, Anna and Wang, Haohuan and Fusai, Niccolo and others},
  journal={arXiv preprint arXiv:2502.19417},
  year={2025}
}

@article{yao2023tree,
  title={Tree of thoughts: Deliberate problem solving with large language models},
  author={Yao, Shunyu and Yu, Dian and Zhao, Jeffrey and Shafran, Izhak and Griffiths, Tom and Cao, Yuan and Narasimhan, Karthik},
  journal={Proc. of the Advances in neural information processing systems (NeurIPS)},
  volume={36},
  pages={11809--11822},
  year={2023}
}

@inproceedings{
  zhao2023large,
  title={Large Language Models as Commonsense Knowledge for Large-Scale Task Planning},
  author={Zirui Zhao and Wee Sun Lee and David Hsu},
  booktitle={Proc. of the Thirty-seventh Conference on Neural Information Processing Systems (NeurIPS)},
  year={2023},
  url={https://openreview.net/forum?id=Wjp1AYB8lH}
}

@inproceedings{
zhang2024restmcts,
title={Re{ST}-{MCTS}*: {LLM} Self-Training via Process Reward Guided Tree Search},
author={Dan Zhang and Sining Zhoubian and Ziniu Hu and Yisong Yue and Yuxiao Dong and Jie Tang},
booktitle={Proc. of the Thirty-eighth Annual Conference on Neural Information Processing Systems (NeurIPS)},
year={2024},
url={https://openreview.net/forum?id=8rcFOqEud5}
}

@misc{
gao2025interpretable,
title={Interpretable Contrastive Monte Carlo Tree Search Reasoning},
author={Zitian Gao and Boye Niu and Xuzheng He and Haotian Xu and Hongzhang Liu and Aiwei Liu and Xuming Hu and Lijie Wen},
year={2025},
url={https://openreview.net/forum?id=F4f1afsm3R}
}

@inproceedings{chi-etal-2025-thoughtsculpt,
    title = "{T}hought{S}culpt: Reasoning with Intermediate Revision and Search",
    author = "Chi, Yizhou  and
      Yang, Kevin  and
      Klein, Dan",
    editor = "Chiruzzo, Luis  and
      Ritter, Alan  and
      Wang, Lu",
    booktitle = "Proc. of the Findings of the Association for Computational Linguistics (NAACL)",
    month = apr,
    year = "2025",
    address = "Albuquerque, New Mexico",
    publisher = "Association for Computational Linguistics",
    url = "https://aclanthology.org/2025.findings-naacl.428/",
    doi = "10.18653/v1/2025.findings-naacl.428",
    pages = "7685--7711",
    ISBN = "979-8-89176-195-7",
}

@inproceedings{
liu2024dont,
title={Don't throw away your value model! Generating more preferable text with Value-Guided Monte-Carlo Tree Search decoding},
author={Jiacheng Liu and Andrew Cohen and Ramakanth Pasunuru and Yejin Choi and Hannaneh Hajishirzi and Asli Celikyilmaz},
booktitle={Proc. of the First Conference on Language Modeling (CoLM)},
year={2024},
url={https://openreview.net/forum?id=kh9Zt2Ldmn}
}

@inproceedings{
liu2023reflect,
title={{REFLECT}: Summarizing Robot Experiences for Failure Explanation and Correction},
author={Zeyi Liu and Arpit Bahety and Shuran Song},
booktitle={Proc. of the 7th Annual Conference on Robot Learning},
year={2023},
url={https://openreview.net/forum?id=8yTS_nAILxt}
}

@article{diehl2022did,
  title={Why did i fail? a causal-based method to find explanations for robot failures},
  author={Diehl, Maximilian and Ramirez-Amaro, Karinne},
  journal={IEEE Robotics and Automation Letters},
  volume={7},
  number={4},
  pages={8925--8932},
  year={2022},
  publisher={IEEE}
}

@ARTICLE{8626460,
  author={Choi, Sungjoon and Lee, Kyungjae and Oh, Songhwai},
  journal={IEEE Transactions on Robotics}, 
  title={Robust Learning From Demonstrations With Mixed Qualities Using Leveraged Gaussian Processes}, 
  year={2019},
  volume={35},
  number={3},
  pages={564-576},
  doi={10.1109/TRO.2019.2891173}}

@article{beyer2024paligemma,
  title={Paligemma: A versatile 3b vlm for transfer},
  author={Beyer, Lucas and Steiner, Andreas and Pinto, Andr{\'e} Susano and Kolesnikov, Alexander and Wang, Xiao and Salz, Daniel and Neumann, Maxim and Alabdulmohsin, Ibrahim and Tschannen, Michael and Bugliarello, Emanuele and others},
  journal={arXiv preprint arXiv:2407.07726},
  year={2024}
}

@inproceedings{
hu2022lora,
title={Lo{RA}: Low-Rank Adaptation of Large Language Models},
author={Edward J Hu and yelong shen and Phillip Wallis and Zeyuan Allen-Zhu and Yuanzhi Li and Shean Wang and Lu Wang and Weizhu Chen},
booktitle={Proc. of the International Conference on Learning Representations (ICLR)},
year={2022},
url={https://openreview.net/forum?id=nZeVKeeFYf9}
}

@inproceedings{liu2024grounding,
  title={Grounding dino: Marrying dino with grounded pre-training for open-set object detection},
  author={Liu, Shilong and Zeng, Zhaoyang and Ren, Tianhe and Li, Feng and Zhang, Hao and Yang, Jie and Jiang, Qing and Li, Chunyuan and Yang, Jianwei and Su, Hang and others},
  booktitle={Proc. of the European conference on computer vision (ECCV)},
  pages={38--55},
  year={2024},
  organization={Springer}
}

@article{hurst2024gpt,
  title={Gpt-4o system card},
  author={Hurst, Aaron and Lerer, Adam and Goucher, Adam P and Perelman, Adam and Ramesh, Aditya and Clark, Aidan and Ostrow, AJ and Welihinda, Akila and Hayes, Alan and Radford, Alec and others},
  journal={arXiv preprint arXiv:2410.21276},
  year={2024}
}

@article{comanici2025gemini,
  title={Gemini 2.5: Pushing the frontier with advanced reasoning, multimodality, long context, and next generation agentic capabilities},
  author={Comanici, Gheorghe and Bieber, Eric and Schaekermann, Mike and Pasupat, Ice and Sachdeva, Noveen and Dhillon, Inderjit and Blistein, Marcel and Ram, Ori and Zhang, Dan and Rosen, Evan and others},
  journal={arXiv preprint arXiv:2507.06261},
  year={2025}
}

@inproceedings{
li2025hamster,
title={{HAMSTER}: Hierarchical Action Models for Open-World Robot Manipulation},
author={Yi Li and Yuquan Deng and Jesse Zhang and Joel Jang and Marius Memmel and Caelan Reed Garrett and Fabio Ramos and Dieter Fox and Anqi Li and Abhishek Gupta and Ankit Goyal},
booktitle={Proc. of the  Thirteenth International Conference on Learning Representations (ICLR)},
year={2025},
url={https://openreview.net/forum?id=h7aQxzKbq6}
}

@article{sutton1999between,
  title={Between MDPs and semi-MDPs: A framework for temporal abstraction in reinforcement learning},
  author={Sutton, Richard S and Precup, Doina and Singh, Satinder},
  journal={Artificial intelligence},
  volume={112},
  number={1-2},
  pages={181--211},
  year={1999},
  publisher={Elsevier}
}

@inproceedings{
ichter2022do,
title={Do As I Can, Not As I Say: Grounding Language in Robotic Affordances},
author={Brian Ichter and Anthony Brohan and Yevgen Chebotar and Chelsea Finn and Karol Hausman and Alexander Herzog and Daniel Ho and Julian Ibarz and Alex Irpan and Eric Jang and Ryan Julian and Dmitry Kalashnikov and Sergey Levine and Yao Lu and Carolina Parada and Kanishka Rao and Pierre Sermanet and Alexander T Toshev and Vincent Vanhoucke and Fei Xia and Ted Xiao and Peng Xu and Mengyuan Yan and Noah Brown and Michael Ahn and Omar Cortes and Nicolas Sievers and Clayton Tan and Sichun Xu and Diego Reyes and Jarek Rettinghouse and Jornell Quiambao and Peter Pastor and Linda Luu and Kuang-Huei Lee and Yuheng Kuang and Sally Jesmonth and Kyle Jeffrey and Rosario Jauregui Ruano and Jasmine Hsu and Keerthana Gopalakrishnan and Byron David and Andy Zeng and Chuyuan Kelly Fu},
booktitle={Proc. of the 6th Annual Conference on Robot Learning (CoRL)},
year={2022},
url={https://openreview.net/forum?id=bdHkMjBJG_w}
}

@article{labbe2020monte,
  title={Monte-carlo tree search for efficient visually guided rearrangement planning},
  author={Labb{\'e}, Yann and Zagoruyko, Sergey and Kalevatykh, Igor and Laptev, Ivan and Carpentier, Justin and Aubry, Mathieu and Sivic, Josef},
  journal={IEEE Robotics and Automation Letters},
  volume={5},
  number={2},
  pages={3715--3722},
  year={2020},
  publisher={IEEE}
}

@inproceedings{
kostrikov2022offline,
title={Offline Reinforcement Learning with Implicit Q-Learning},
author={Ilya Kostrikov and Ashvin Nair and Sergey Levine},
booktitle={Proc. of the International Conference on Learning Representations (ICLR)},
year={2022},
url={https://openreview.net/forum?id=68n2s9ZJWF8}
}

@article{margellos2011hamilton,
  title={Hamilton--Jacobi formulation for reach--avoid differential games},
  author={Margellos, Kostas and Lygeros, John},
  journal={IEEE Transactions on automatic control},
  volume={56},
  number={8},
  pages={1849--1861},
  year={2011},
  publisher={IEEE}
}

@inproceedings{fisac2015reach,
  title={Reach-avoid problems with time-varying dynamics, targets and constraints},
  author={Fisac, Jaime F and Chen, Mo and Tomlin, Claire J and Sastry, S Shankar},
  booktitle={Proc. of the 18th international conference on hybrid systems: computation and control},
  pages={11--20},
  year={2015}
}

@inproceedings{
agia2024unpacking,
title={Unpacking Failure Modes of Generative Policies: Runtime Monitoring of Consistency and Progress},
author={Christopher Agia and Rohan Sinha and Jingyun Yang and Ziang Cao and Rika Antonova and Marco Pavone and Jeannette Bohg},
booktitle={Proc. of the 8th Annual Conference on Robot Learning (CoRL)},
year={2024},
url={https://openreview.net/forum?id=yqLFb0RnDW}
}

@inproceedings{wortsman2022model,
  title={Model soups: averaging weights of multiple fine-tuned models improves accuracy without increasing inference time},
  author={Wortsman, Mitchell and Ilharco, Gabriel and Gadre, Samir Ya and Roelofs, Rebecca and Gontijo-Lopes, Raphael and Morcos, Ari S and Namkoong, Hongseok and Farhadi, Ali and Carmon, Yair and Kornblith, Simon and others},
  booktitle={Proc. of the International conference on machine learning (ICML)},
  pages={23965--23998},
  year={2022},
  organization={PMLR}
}

@article{belkhale2023data,
  title={Data quality in imitation learning},
  author={Belkhale, Suneel and Cui, Yuchen and Sadigh, Dorsa},
  journal={Proc. of the Advances in neural information processing systems (NeurIPS)},
  volume={36},
  pages={80375--80395},
  year={2023}
}

@book{sutton1998introduction,
  title={Introduction to reinforcement learning},
  author={Sutton, Richard S and Barto, Andrew G and others},
  volume={135},
  year={1998},
  publisher={MIT press Cambridge}
}

@article{kaelbling1996reinforcement,
  title={Reinforcement learning: A survey},
  author={Kaelbling, Leslie Pack and Littman, Michael L and Moore, Andrew W},
  journal={Journal of artificial intelligence research},
  volume={4},
  pages={237--285},
  year={1996}
}

@inproceedings{
li2024evaluating,
title={Evaluating Real-World Robot Manipulation Policies in Simulation},
author={Xuanlin Li and Kyle Hsu and Jiayuan Gu and Oier Mees and Karl Pertsch and Homer Rich Walke and Chuyuan Fu and Ishikaa Lunawat and Isabel Sieh and Sean Kirmani and Sergey Levine and Jiajun Wu and Chelsea Finn and Hao Su and Quan Vuong and Ted Xiao},
booktitle={Proc. of the 8th Annual Conference on Robot Learning (CoRL)},
year={2024},
url={https://openreview.net/forum?id=LZh48DTg71}
}

@inproceedings{
walke2023bridgedata,
title={BridgeData V2: A Dataset for Robot Learning at Scale},
author={Homer Rich Walke and Kevin Black and Tony Z. Zhao and Quan Vuong and Chongyi Zheng and Philippe Hansen-Estruch and Andre Wang He and Vivek Myers and Moo Jin Kim and Max Du and Abraham Lee and Kuan Fang and Chelsea Finn and Sergey Levine},
booktitle={Proc. of the 7th Annual Conference on Robot Learning (CoRL)},
year={2023},
url={https://openreview.net/forum?id=f55MlAT1Lu}
}

@inproceedings{zhao2023learning,
  title={Learning fine-grained bimanual manipulation with low-cost hardware},
  author={Zhao, Tony Z and Kumar, Vikash and Levine, Sergey and Finn, Chelsea},
  booktitle={Proc. of the Robotics: Science and System (RSS)},
  year={2023}
}

@inproceedings{
lipman2023flow,
title={Flow Matching for Generative Modeling},
author={Yaron Lipman and Ricky T. Q. Chen and Heli Ben-Hamu and Maximilian Nickel and Matthew Le},
booktitle={Proc. of the Eleventh International Conference on Learning Representations (ICLR) },
year={2023},
url={https://openreview.net/forum?id=PqvMRDCJT9t}
}

@article{liang2022code,
  title={Code as policies: Language model programs for embodied control},
  author={Liang, Jacky and Huang, Wenlong and Xia, Fei and Xu, Peng and Hausman, Karol and Ichter, Brian and Florence, Pete and Zeng, Andy},
  journal={arXiv preprint arXiv:2209.07753},
  year={2022}
}

@INPROCEEDINGS{10161317,
  author={Singh, Ishika and Blukis, Valts and Mousavian, Arsalan and Goyal, Ankit and Xu, Danfei and Tremblay, Jonathan and Fox, Dieter and Thomason, Jesse and Garg, Animesh},
  booktitle={Proc. of the 2023 IEEE International Conference on Robotics and Automation (ICRA)}, 
  title={ProgPrompt: Generating Situated Robot Task Plans using Large Language Models}, 
  year={2023},
  volume={},
  number={},
  pages={11523-11530},
  doi={10.1109/ICRA48891.2023.10161317}}

@inproceedings{yao2023react,
  title={React: Synergizing reasoning and acting in language models},
  author={Yao, Shunyu and Zhao, Jeffrey and Yu, Dian and Du, Nan and Shafran, Izhak and Narasimhan, Karthik and Cao, Yuan},
  booktitle={Proc. of the International Conference on Learning Representations (ICLR)},
  year={2023}
}

@inproceedings{
hejna2024remix,
title={ReMix: Optimizing Data Mixtures for Large Scale Imitation Learning},
author={Joey Hejna and Chethan Anand Bhateja and Yichen Jiang and Karl Pertsch and Dorsa Sadigh},
booktitle={Proc. of the 8th Annual Conference on Robot Learning (CoRL)},
year={2024},
url={https://openreview.net/forum?id=fIj88Tn3fc}
}

@inproceedings{ni2024grid,
  title={Grid: Scene-graph-based instruction-driven robotic task planning},
  author={Ni, Zhe and Deng, Xiaoxin and Tai, Cong and Zhu, Xinyue and Xie, Qinghongbing and Huang, Weihang and Wu, Xiang and Zeng, Long},
  booktitle={Proc. of the 2024 IEEE/RSJ International Conference on Intelligent Robots and Systems (IROS)},
  pages={13765--13772},
  year={2024},
  organization={IEEE}
}

@InProceedings{rana23a,
  title = 	 {SayPlan: Grounding Large Language Models using 3D Scene Graphs for Scalable Robot Task Planning},
  author =       {Rana, Krishan and Haviland, Jesse and Garg, Sourav and Abou-Chakra, Jad and Reid, Ian and Suenderhauf, Niko},
  booktitle = 	 {Proc. of The 7th Conference on Robot Learning (CoRL)},
  pages = 	 {23--72},
  year = 	 {2023},
}

@book{kahneman2011thinking,
  title={Thinking, fast and slow},
  author={Kahneman, Daniel},
  year={2011},
  publisher={macmillan}
}

@inproceedings{
hoang2024sprinql,
title={{SPRINQL}: Sub-optimal Demonstrations driven Offline Imitation Learning},
author={Huy Hoang and Tien Anh Mai and Pradeep Varakantham},
booktitle={Proc. of the Thirty-eighth Annual Conference on Neural Information Processing Systems (NeuRIPS)},
year={2024},
url={https://openreview.net/forum?id=uDD44NROOt}
}

@inproceedings{freitag2017beam,
    title = "Beam Search Strategies for Neural Machine Translation",
    author = "Freitag, Markus  and
      Al-Onaizan, Yaser",
    editor = "Luong, Thang  and
      Birch, Alexandra  and
      Neubig, Graham  and
      Finch, Andrew",
    booktitle = "Proc. of the First Workshop on Neural Machine Translation",
    month = aug,
    year = "2017",
    publisher = "Association for Computational Linguistics",
    pages = "56--60",
}

@ARTICLE{9140424,
  author={Larsson, Daniel T. and Maity, Dipankar and Tsiotras, Panagiotis},
  journal={IEEE Transactions on Robotics}, 
  title={Q-Tree Search: An Information-Theoretic Approach Toward Hierarchical Abstractions for Agents With Computational Limitations}, 
  year={2020},
  volume={36},
  number={6},
  pages={1669-1685},
  doi={10.1109/TRO.2020.3003219}}

@ARTICLE{4084578,
  author={Yershova, Anna and LaValle, Steven M.},
  journal={IEEE Transactions on Robotics}, 
  title={Improving Motion-Planning Algorithms by Efficient Nearest-Neighbor Searching}, 
  year={2007},
  volume={23},
  number={1},
  pages={151-157},
  doi={10.1109/TRO.2006.886840}}

@ARTICLE{1492476,
  author={Plaku, E. and Bekris, K.E. and Chen, B.Y. and Ladd, A.M. and Kavraki, L.E.},
  journal={IEEE Transactions on Robotics}, 
  title={Sampling-based roadmap of trees for parallel motion planning}, 
  year={2005},
  volume={21},
  number={4},
  pages={597-608},
  doi={10.1109/TRO.2005.847599}}

@ARTICLE{11024207,
  author={Liu, Chuhao and Qiao, Zhijian and Shi, Jieqi and Wang, Ke and Liu, Peize and Shen, Shaojie},
  journal={IEEE Transactions on Robotics}, 
  title={SG-Reg: Generalizable and Efficient Scene Graph Registration}, 
  year={2025},
  volume={41},
  number={},
  pages={3870-3889},
  doi={10.1109/TRO.2025.3577020}}

@INPROCEEDINGS{6386109,
  author={Todorov, Emanuel and Erez, Tom and Tassa, Yuval},
  booktitle={Proc. of the IEEE/RSJ International Conference on Intelligent Robots and Systems (IROS)}, 
  title={MuJoCo: A physics engine for model-based control}, 
  year={2012},
  volume={},
  number={},
  pages={5026-5033},
  doi={10.1109/IROS.2012.6386109}}
